%% file: root.tex
\documentclass[iicol,pdflatex]{sn-jnl}

\input{preamble}

\def\tfs{7}
\def\bfs{10}

\begin{document}
\title[Article Title]{Reinforcement Learning based Autonomous Multi-Rotor Landing on Moving Platforms}
\author*[1,2]{\fnm{Pascal} \sur{Goldschmid}}\email{pascal.goldschmid@ifr.uni-stuttgart.de}
\author[1,2]{\fnm{Aamir} \sur{Ahmad}}\email{aamir.ahmad@ifr.uni-stuttgart.de}
\affil[1]{\orgname{University of Stuttgart}, \orgaddress{\postcode{70569} \city{Stuttgart}, \country{Germany}}}
\affil[2]{\orgname{Max Planck Institute for Intelligent Systems}, \orgaddress{\postcode{72076} \city{T\"ubingen}, \country{Germany}}}
\maketitle 

\input{./01_abstract.tex}
\input{./02_introduction.tex}
\input{./03_methodology.tex}

\input{./04_implementation.tex}
\input{./05_experiments.tex}

\bibliographystyle{sn-basic.bst}
\bibliography{new_bib}
\end{document}

%% file: preamble.tex
\usepackage{graphicx}%
\usepackage{multirow}%
\usepackage{amsmath}
\usepackage{amssymb}
\usepackage{amsfonts}
\usepackage{amsthm}%
\usepackage{mathrsfs}%
\usepackage[title]{appendix}%
\usepackage{xcolor}%
\usepackage{textcomp}%
\usepackage{manyfoot}%
\usepackage{booktabs}%
\usepackage{algorithm}%
\usepackage{algorithmicx}%
\usepackage{algpseudocode}%
\usepackage{listings}%
\usepackage{hyperref}
\usepackage {abstract}
\usepackage{siunitx}
\DeclareMathOperator*{\argmax}{\arg\!\max}
\usepackage{makecell}
\usepackage{caption}
\usepackage[authoryear]{natbib}

%% file: 01_abstract.tex
\begin{abstract}\vphantom{M}\\Multi-rotor UAVs suffer from a restricted range and flight duration due to limited battery capacity. Autonomous landing on a 2D moving platform offers the possibility to replenish batteries and offload data, thus increasing the utility of the vehicle. Classical approaches rely on accurate, complex and difficult-to-derive models of the vehicle and the environment. Reinforcement learning (RL) provides an attractive alternative due to its ability to learn a suitable control policy exclusively from data during a training procedure. However, current methods require several hours to train, have limited success rates and depend on hyperparameters that need to be tuned by trial-and-error. \\
We address all these issues in this work. First, we decompose the landing procedure into a sequence of simpler, but similar learning tasks. This is enabled by applying two instances of the same RL based controller trained for 1D motion for controlling the multi-rotor's movement in both the longitudinal and the lateral directions. Second, we introduce a powerful state space discretization technique that is based on i) kinematic modeling of the moving platform to derive information about the state space topology and ii) structuring the training as a sequential curriculum using transfer learning.  Third, we leverage the kinematics model of the moving platform to also derive interpretable hyperparameters for the training process that ensure sufficient maneuverability of the multi-rotor vehicle. The training is  performed using the tabular RL method \emph{Double Q-Learning}. \\
Through extensive simulations we show that the presented method significantly increases the rate of successful landings,  while requiring less training time compared to other deep RL approaches. \textcolor{black}{Furthermore, for two comparison scenarios it achieves comparable performance than a cascaded PI controller.} Finally, we deploy and demonstrate our algorithm on real hardware. For all evaluation scenarios we provide statistics on the agent's performance. Source code is openly available at \url{https://github.com/robot-perception-group/rl_multi_rotor_landing}.
\end{abstract}

%% file: 02_introduction.tex
\section{Introduction}
Classical control approaches suffer from the fact that accurate models of the plant and the environment are required for the controller design. These models are necessary to consider non-linearities of the plant behavior. However, they are subjected to limitations, e.g., with regard to disturbance rejection \citep{Rodriguez-Ramos2019} and parametric uncertainties \citep{Mo2018}.
To overcome these problems, reinforcement learning (RL) has been studied for robot control in the last decades, including in the context of UAV control \citep{Mo2018}. 
In model-free, (action-) value based RL methods, an approximation of a value function is learned exclusively through interaction with the environment. On this basis, a policy then selects an optimal action while being in a certain state \citep{Sutton2015}. Classical RL approaches such as Q-Learning \citep{Sutton2015} store a tabular representation of that value function approximation. Deep Reinforcement Learning (DRL) methods \citep{Mnih2013}, on the other hand, leverage a deep neural network (DNN) to learn an approximate value function over a continuous state and action space, thus making it a powerful approach for many complex applications. Nevertheless, in both RL and DRL, the training itself is in general sensitive to the formulation of the learning task and in particular to the selection of hyperparameters. This can result in long training times, the necessity to perform a time-consuming hyperparameter search and an unstable convergence behavior. Furthermore, especially in DRL, the neural network acts like a black box, i.e. it is difficult to trace back which training experience caused certain aspects of the learned behaviour.
The purpose of this work is to address the aforementioned issues for the task of landing a multi-rotor UAV on a moving platform. Our RL based method aims at i) achieving a high rate of successful landing attempts, ii) requiring short training time, and iii) providing intrepretable ways to compute hyperparameters necessary for training.

\begin{figure}[!t]
      \centering	
       \includegraphics[width=7.5cm]{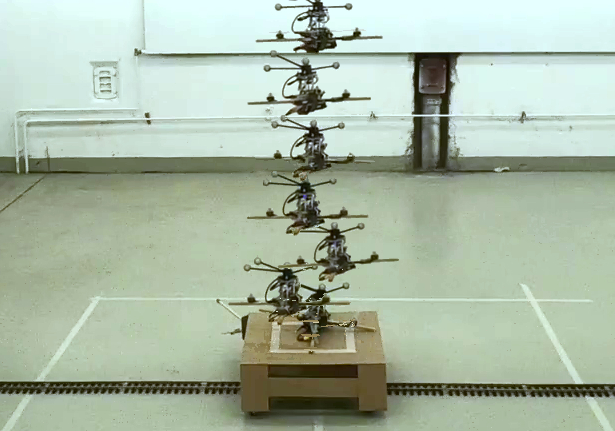}
\caption{Our multi-rotor vehicle landing on a platform moving on rails.}
\label{fig:title_image}
\end{figure}

To achieve these aims, we leverage the tabular off-policy algorithm Double Q-Learning \citep{Hasselt2010}, due to the few required hyperparameters which are the decay factor $\gamma$ and the learning rate $\alpha$. Using a tabular method does not require defining the DNN's  architecture or finding values for additional hyperparameters such as minibatch and buffer size or the number of gradient steps. Furthermore, unlike a NN-based deep learning algorithm, it does not require a powerful GPU for training, making it also suitable for computers with lower performance and less power-consuming.
However, tabular methods only provide discrete state and action spaces and therefore suffer from the ``curse of dimensionality'' \citep{Lampton2009}. Furthermore, the training performance and control performance is influenced by the sampling rate. When the sampling \textcolor{black}{rate} and discretization do not match, the performance of the agent can be low, e. g. due to jittering, where the agent rapidly alternates between discrete states. To solve these problems, our method's \textbf{novel aspects} are outlined below. The first two of these allow us to address the ``curse of dimensionality''  issue by reducing the complexity of the learning task. The third addresses the problem to find a matching discretization and sampling rate.

\begin{enumerate}
 \item Under the assumption of a symmetric UAV and decoupled dynamics, common in literature \citep{Wang2016}, our method is able to control the vehicle's motion in longitudinal and lateral direction with two instances of the \emph{same} RL agent. Thus, the learning task is simplified to 1D movement only. The vertical and yaw movements are controlled by PID controllers. The concept of using independent controllers for different directions of motion is an approach which is often used, e.g. when PID controllers are applied (\citeauthor{Wenzel2011}, \citeyear{Wenzel2011}; \citeauthor{Araar2017}, \citeyear{Araar2017}).
 \item We introduce a novel state-space discretization approach, motivated by the insight that precise knowledge of the relative position and relative velocity between the UAV and the moving platform is required only when touchdown is imminent. To this end, we leverage a multiresolution technique \citep{Lampton2009} and augment it with information about the state space topology, derived from a simple kinematics model of the moving platform. This allows us to restructure the learning task of the 1D movement as a sequence of even simpler problems by means of a sequential curriculum, in which learned action values are transferred to the subsequent curriculum step. Furthermore, the discrete state space allows us to accurately track how often a state has been visited during training.
\item We leverage the discrete action space to ensure sufficient maneuverability of the UAV to follow the platform. To this end, we derive equations that compute the values of hyperparameters, such as the agent frequency and the maximum value of the roll/pitch angle of the UAV. The intention of these equations is twofold. First, they link the derived values to the maneuverability of the UAV in an interpretable way. Second, they ensure that the discretization of the state space matches the agent frequency. The aim is to reduce unwanted side effects resulting from the application of a discrete state space, such as jittering.
\end{enumerate}
Sect.~\ref{sec:related_work} presents related work, followed by Sect.~\ref{sec:Methodology} describing the proposed approach in detail. Sect.~\ref{sec:Implementation} presents the implementation and  experimental setup. The results are discussed in Sect.~\ref{sec:Experiments}, with comments on future work given in Sect. ~\ref{sec:conclusion}.

\section{Related Work} \label{sec:related_work}

\subsection{Classical Control Approaches}
So far, the problem of landing a multi-rotor aerial vehicle has been tackled for different levels of complexity regarding platform movement.
One-dimensional platform movement is considered in  \citep{Hu2015} in the context of maritime applications,  where a platform is oscillating vertically. Two-dimensional platform movement is treated in (\citeauthor{Wenzel2011}, \citeyear{Wenzel2011}; \citeauthor{Ling2014}, \citeyear{Ling2014}; \citeauthor{Gautam2015}, \citeyear{Gautam2015}; \citeauthor{Vlantis2015}, \citeyear{Vlantis2015};  \citeauthor{Araar2017}, \citeyear{Araar2017}; \citeauthor{Borowczyk2017}, \citeyear{Borowczyk2017}; \citeauthor{Falanga2017}, \citeyear{Falanga2017}). Three-dimensional translational movement of the landing platform is covered for docking scenarios involving multi-rotor vehicles in  \citep{Zhong2016} and \citep{Miyazaki2018}. 
Various control techniques have been applied to enable a multi-rotor UAV to land in one of these scenarios. In \citep{Hu2015} an adaptive robust controller is used to control the vehicle during a  descend maneuver onto a vertically oscillating platform while considering the ground effect during tracking of a reference trajectory.
The authors of \citep{Gautam2015} apply guidance laws that are based on missile control principles. \citep{Vlantis2015} uses a model predictive controller to land on an inclined moving platform.  \citep{Falanga2017} relies on a non-linear control approach involving LQR-controllers. However, the most used controller type is a PID controller.
 Reference \citep{Wenzel2011} uses four independent PID controllers to follow the platform that has been identified with visual tracking methods. In \citep{Ling2014}, the landing maneuver onto a maritime vessel is structured into different phases. Rendezvous with the vessel, followed by aquiring a visual fiducial marker and descending onto the ship. During all phases, PID controllers are used to control the UAV.
Perception and relative pose estimation based methods are the focus of \citep{Araar2017}, where again four PID controllers provide the UAV with the required autonomous motion capability. A PID controller is also applied in \citep{Borowczyk2017} to handle the final touchdown on a ground vehicle moving up to $50\si{km/h}$. Also the landing on a 3D moving platform can be solved with PID controllers (\citeauthor{Zhong2016}, \citeyear{Zhong2016}; \citeauthor{Miyazaki2018}, \citeyear{Miyazaki2018}). Although automatic methods for tuning the gains of a PID controller exist, tuning is often a manual, time-consuming procedure. However, learning-based control methods enable obtaining a suitable control policy exclusively from data through interaction with the environment, making it an attractive and superior approach for controlling an UAV to land on a moving platform. Furthermore, methods such as Q-Learning enable the approximation of an optimal action-value function and thus lead to a (near-) optimal action selection policy \citep{Sutton2015}.

\subsection{Learning-based Control Approaches}
The landing problem has been approached with (Deep) Reinforcement Learning for static platforms (\citeauthor{Kooi2021}, \citeyear{Kooi2021}; \citeauthor{Shi2019}, \citeyear{Shi2019}; \citeauthor{Polvara2018}, \citeyear{Polvara2018}; \citeauthor{Polvara2019}, \citeyear{Polvara2019}) and for moving  platforms (\citeauthor{Rodriguez-Ramos2019}, \citeyear{Rodriguez-Ramos2019}; \citeauthor{Lee2018} \citeyear{Lee2018}).  
The authors of \citep{Kooi2021} accelerate training by means of a curriculum, where a policy is learned using Proximal Policy Optimization (PPO) to land on an inclined, static platform. Their curriculum is  tailored, involving several  hyperparameters, whereas in this work we present a structured approach for deriving a curriculum for different scenarios. 
In \citep{Polvara2018} the authors assign different control tasks to different agents. A DQN-agent aligns a quadcopter with a marker on the ground, whereas a second agent commands a descending maneuver before a closed-loop controller handles the touchdown. \citep{Polvara2019} follows a similar approach but allows small pitch and roll maneuvers of the platform.  Our approach also leverages separate RL agents, but for controlling the longitudinal and lateral motion. 
The authors of \citep{Shi2019} present a deep-learning-based robust non-linear controller to increase the precision of landing and close ground trajectory following. A nominal dynamics model is combined with a DNN to learn the ground effect on aerodynamics and UAV dynamics. 
\citep{Lee2018} presents an actor-critic approach for landing using continuous state and action spaces to control the roll and pitch angle of the drone, but no statistics on its performance.
In \citep{Rodriguez-Ramos2019}, a DRL-framework is introduced for training in simulation and evaluation in real world. The Deep Deterministic Policy Gradient (DDPG) algorithm, an actor-critic approach, is used to command the movement of a multi-rotor UAV in longitudinal and lateral direction with continuous states and actions. Detailed data about the agent's success rate in landing in simulation is provided. We use this work as a baseline method and show how we outperform it, providing also statistics about our agent's performance in the real world. However, the baseline method does not provide any systematic and explainable way for deriving hyperparameters used for the learning problem. For our method, we present equations that link  values of hyperparameters to problem properties such as the state space discretization and maneuverability of the UAV in an intuitively understandable way. 

%% file: 03_methodology.tex
\section{Methodology}\label{sec:Methodology}
\subsection{Preliminaries}
\subsubsection{Problem Statement}
The goal is to develop a RL based approach that provides a high rate of successful UAV landings on a horizontally moving platform. A landing trial is considered successful if the UAV touches down on the surface of the moving platform.  The approach should require less training time than the baseline method and generalize well enough to be deployed on real hardware. Furthermore, required hyperparameters should be determined in an interpretable way.

\subsubsection{Base Notations} \label{sec:prelim_notations}
\begin{figure}[thpb]
      \centering
       \includegraphics[scale=0.47]{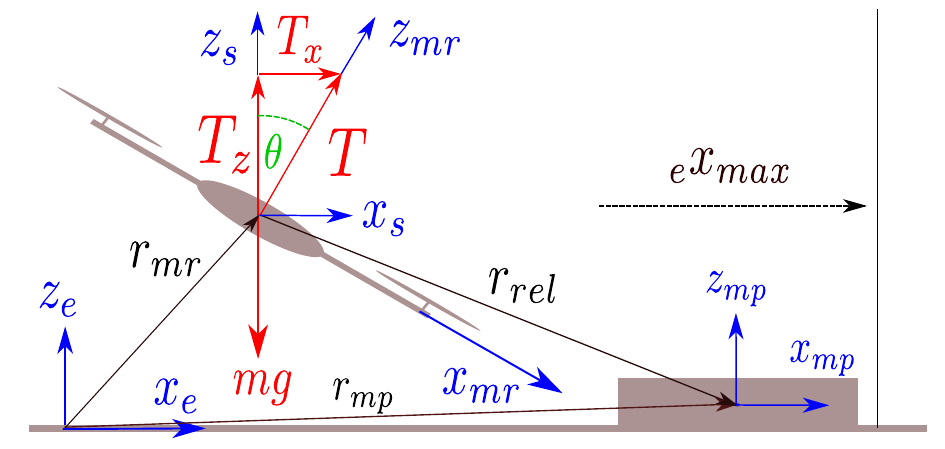}			
\caption{Coordinate frames (blue, $e$: earth frame, $mp$: body-fixed frame of moving platform, $mr$: body-fixed frame of multi-rotor UAV, $s$: stability frame), forces (red) and attitude angle (green) associated with 1D motion in longitudinal direction.}
\label{fig:model}
 \end{figure}
We first define a fly zone $\mathcal{F}=[-x_{max},x_{max}]\times [-y_{max},y_{max}] \times [0,z_{max}]\subset\mathbb{R}^3[\si{m}]$ in which the motion of the multi-rotor UAV $mr$ and the moving platform $mp$ are considered, as is illustrated by Fig.~\ref{fig:model}. Furthermore, for the goal of landing on the moving platform, the RL agent controlling the multi-rotor vehicle has to consider only the relative translational and rotational motion. This is why we express the relative dynamics and kinematics in the stability frame $s$, to be able to formulate the RL problem from the UAV's and thus the agent's point of view. 
For this purpose we first denote the  Euler angles roll, pitch and yaw in the stability frame with $_s\boldsymbol\varphi= (\phi,\theta,\psi)^T$. Each landing trial will begin with the UAV being in hover state, leading to the following initial conditions for rotational movement $ {}_s\dot{\boldsymbol\varphi}_{mr,0} = {}_s\dot{\boldsymbol\varphi}_{mp,0} = \mathbf{0} \si{rad/s} $, $ {}_s\boldsymbol\varphi_{mr,0} = {}_s\boldsymbol\varphi_{mp,0} = \mathbf{0} \si{rad}$. 
The initial conditions for the multi-rotor vehicle's translational motion are expressed in the earth-fixed frame $e$, it is $_e\dot{\mathbf{r}}_{mr,0} =\mathbf{0}\si{m/s}$ and $_e\mathbf{r}_{mr,0} \in \mathcal{F}$. Transforming the translational initial conditions of both, multi-rotor vehicle and moving platform into the stability frame allows to fully express the relative motion as the second order differential equations

\begin{small}
\begin{equation}
_{s}\ddot{\boldsymbol\varphi}_{rel} = \mathbf{0} - {}_s\ddot{\boldsymbol\varphi}_{mr} ~~~ \textrm{and}  ~~~ {}_{s}\ddot{\mathbf{r}}_{rel} =  {}_s\ddot{\mathbf{r}}_{mp} - {}_s\ddot{\mathbf{r}}_{mr}.
\label{eq:non_linear_rel_motion}
\end{equation}
\end{small}

For the remainder of this work we specify the following values that will serve as continuous observations of the environment (see Sec.~\ref{sec:discrete_state_space} and Fig.~\ref{fig:sim_framework})

\begin{small}
\begin{align}
\mathbf{p}_c &= [p_{c,x},p_{c,y},p_{c,z}]^T = {}_s\mathbf{r}_{rel} \label{eq:cont_obs_p_cx}\\
\mathbf{v}_c &= [v_{c,x},v_{c,y},v_{c,z}]^T = {}_s\dot{\mathbf{r}}_{rel}\\
\mathbf{a}_c &= [a_{c,x},a_{c,y},a_{c,z}]^T = {}_s\ddot{\mathbf{r}}_{rel}\\
\boldsymbol\varphi_c &= [\phi_{rel},\theta_{rel},\psi_{rel}]^T = {}_{s}\boldsymbol\varphi_{rel}.\label{eq:cont_obs_phi_rel}
\end{align}
\end{small}

$\mathbf{p}_c$ denotes the relative position, $\mathbf{v}_c$  the relative velocity, $\mathbf{a}_c$  the relative acceleration   and $\boldsymbol\varphi_c$ the relative orientation between vehicle and platform,  expressed in the stability frame.

\subsection{Motion of the Landing Platform}
We introduce the rectilinear periodic movement (RPM) of the platform that is applied  during training. With the initial conditions $_e\dot{\mathbf{r}}_{mp,0} = \mathbf{0}\si{m/s}$ and $_e\mathbf{r}_{mp,0} = \mathbf{0}\si{m}$, the translational motion of the platform is expressed as a second order differential equation in the earth-fixed frame \textit{e}.

\begin{small}
\begin{equation}
\begin{split}
    _e\ddot{\mathbf{r}}_{mp} &=-\left(v_{mp}^2/r_{mp}\right)\left[\sin(\omega_{mp} t),0,0\right]^T, \\
     \omega_{mp} &= v_{mp}/r_{mp}\\
\end{split} 
\label{eq:platform_acceleration_rpm}
\end{equation}
\end{small}

\noindent where $r_{mp}$ denotes the maximum amplitude of the trajectory, $v_{mp}$ the maximum platform velocity and $\omega_{mp}$ the angular frequency of the rectilinear movement. The platform  is not subjected to any rotational movement. As a consequence, the maximum acceleration of the platform that is required as part of the hyperparameter determination in Sec.~\ref{sec:hyperparemter_estimation} is

\begin{small}
\begin{equation}
a_{mp,max} = v_{mp}^2/r_{mp}.
\label{eq:a_mpmax}
\end{equation}
\end{small}

\subsection{Basis Learning Task} \label{sec:basis_learning_task}

 \subsubsection{Dynamic Decoupling} \label{sec:basis_learning_task_preliminaries}
 The nonlinear dynamics of a multi-rotor vehicle such as a quadcopter can be greatly simplified when the following assumptions are applied.  i) Linearization around hover flight, ii) small angles iii) a rigid, symmetric vehicle body, iv) no aerodynamic effects \citep{Wang2016}. 
 Under these conditions, the axes of motion are decoupled. Thus, four individual controllers can be used to independently control the  movement in longitudinal, lateral,  vertical direction and  around the vertical axis. For this purpose, low-level controllers track a setpoint $\theta_{ref}$  for the pitch angle (longitudinal motion), $\phi_{ref}$ for the roll angle  (lateral motion), $T_{ref}$ for the total thrust (vertical motion) and $\psi_{rel}$ for the yaw angle (rotation around vertical axis).  We leverage this fact in our controller structure as is further illustrated in Sect.~\ref{sec:Implementation}. 
Furthermore, this enables us to introduce a learning task for longitudinal 1D motion only.  Its purpose is to learn a policy for the pitch angle setpoint $\theta_{ref}$ that induces a 1D movement in longitudinal direction, allowing the vehicle to center over the platform moving in longitudinal direction as well.  After the training, we then apply a second instance of the trained agent for controlling the lateral motion, where it produces set point values for the roll angle $\phi_{ref}$. This way, full 2D motion capability is achieved. We use the basis learning task to compose a curriculum later in Sect.~\ref{sec:curriculum_discretization}. PID controllers ensure that $\phi_{rel} =0\si{rad},\psi_{rel} = 0\si{rad}$ and $ v_{z_{rel}} = \text{const} < 0\si{m/s}$  during training.

\subsubsection{Markov Decision Process}
For the RL task formulation we consider the finite, discrete Markov Decision Process \citep{Sutton2015}.  
The goal of the agent is to learn the optimal action value function $Q^*$ to obtain the optimal policy $\pi^* = \argmax_a Q^*(s,a)$, maximizing the sum of discounted rewards $\Sigma^{T-t-1}_{k = 0}\gamma^{k}r_{t+k+1}$.

\subsubsection{Discrete Action Space}
We choose a discrete action space comprising three actions, namely \textit{increasing pitch}, \textit{decreasing pitch} and \textit{do nothing}, denoted by

\begin{small}
\begin{align}
\mathbb{A_d} = \left[ \Delta\theta^{+},\Delta\theta^{-}, -\right].
\label{eq:discrete_action_space}
\end{align}
\end{small}

The pitch angle increment is defined by

\begin{small}
\begin{equation}
\Delta\theta = \frac{\theta_{max}}{n_{\theta}}, n_{\theta}\in \mathbb{N}^+,
\label{eq:normalized_pitch_set}
\end{equation}
\end{small}

\noindent where $\theta_{max}$ denotes the maximum pitch angle and  $n_{\theta}$  the number of intervals intersecting the range $[0,\theta_{max}]$.
The set of possible pitch angles that can be taken by the multi-rotor vehicle is then 

\begin{small}
\begin{align}
\begin{split}
\Theta &= \left\lbrace -\theta_{max} + i_{\theta}\Delta\theta\lvert i_{\theta} \in \left\lbrace 0,\ldots,2n_{\theta}\right\rbrace\right\rbrace,
\end{split}
\end{align}
\end{small}

\noindent where $i_{\theta}$ is used to select a specific element.

\subsubsection{Discrete State Space} \label{sec:discrete_state_space}
For the discrete state space we first scale and clip the continuous observations of the environment that are associated with motion in the longitudinal direction to a value range of $[-1,1]$. For this purpose, we use the function $\text{clip}(x,x_{min},x_{max})$ that clips the value of $x$ to the range of $[x_{min},x_{max}]$.

\begin{small}
\begin{align}
p_x &= \text{clip}(p_{c,x}/p_{max},-1,1),\label{eq:xyz_c_x}
\\
 v_x& = \text{clip}(v_{c,x}/v_{max},-1,1),  \\
a_x &= \text{clip}(a_{c,x}/a_{max},-1,1),
\label{eq:xyz_c_z}
\end{align}
\end{small}

\noindent where $p_{max},v_{max},a_{max}$ are the values used for the normalization of the observations. The reason for the clipping is that our state space discretization technique, a crucial part of the sequential curriculum, is derived assuming a worst case scenario for the platform movement in which the multi-rotor vehicle is hovering (see Sect.~\ref{sec:curriculum_discretization}) and where scaling an observation with its maximum values, $p_{max}, v_{max}$ and $a_{max}$, would constitute a normalization. However, once the multi-rotor vehicle starts moving too, the scaled observation values could exceed the value range of $[-1,1]$. Clipping  allows the application of the discretization technique also for a moving multi-rotor vehicle.
Next, we define a general discretization function  $d(x,x_1,x_2)$ that can be used to map a continuous observation of the environment to a discrete state value.

\begin{small}
\begin{equation}
d(x,x_1,x_2) =
\begin{cases}
& 0 \text{ if } x\in [-x_2,-x_1)\\
& 1 \text{ if } x\in [-x_1,x_1] \\
& 2 \text{ if } x\in (x_1,x_2]
\end{cases}
\label{eq:mapping}
\end{equation}
\end{small}

We apply \eqref{eq:mapping} to the normalized observations \eqref{eq:xyz_c_x}-\eqref{eq:xyz_c_z} to determine the discrete state  $s =(p_d, v_d,a_d,i_{\theta}) \in \mathbb{S}$, where $i_{\theta} \in \left\lbrace 0,\ldots,2n_{\theta}\right\rbrace$, $\mathbb{S}= \mathbb{N}_0^{ 3\times  3\times 3\times 2n_{\theta}+1}$ and

\begin{small}
\begin{align}
p_d &= d\left(p_x,p_{goal},p_{lim}\right),\label{eq:p_d}\\
v_d &= d\left(v_x,v_{goal},v_{lim}\right), \label{eq:v_d}\\
 a_d &= d\left(a_x,a_{goal},a_{lim}\right).\label{eq:a_d}
\end{align}
\end{small}

In \eqref{eq:p_d}-\eqref{eq:a_d}, the normalized values $\pm p_{goal}\pm ,v_{goal}, \pm a_{goal}$ define the boundaries of the discrete states the agent should learn to reach whereas the normalized values of $\pm p_{lim}\pm ,v_{lim}, \pm a_{lim}$ denote the limits in the observations the agent should learn not to exceed when controlling the multi-rotor vehicle.

\subsubsection{Goal State}
We define the goal state  $s^* \in \mathbb{S}$

\begin{small}
\begin{equation}
s^* = \left\lbrace 1,1,*,*\right\rbrace.
\label{eq:goal_state}
\end{equation}
\end{small}

This means the goal state is reached if $-p_{goal}\leq p_x\leq p_{goal}$ and $-v_{goal}\leq v_x\leq v_{goal}$  regardless of the values of $a_d$ and $i_{\theta}$. 

\subsubsection{Reward Function}\label{sec:reward_function}
Our reward function $r_t$, inspired by a shaping approach \citep{Rodriguez-Ramos2018} is given as 

\begin{small}
\begin{align}
r_t =  r_p + r_v + r_{\theta} + r_{dur} + r_{term},
\label{eq:reward}
\end{align}
\end{small}

\noindent where $r_{term} = r_{suc}$ if $s=s^*$ and $r_{term} = r_{fail}$ if  $|p_x|>p_{lim}$ or if the maximum episode duration $t_{max}$ is reached. In all other cases, $r_{term} = 0$. $r_{suc}$ and $r_{fail}$ will be defined at the end of this section. Positive rewards $r_p$, $r_v$ and $r_{\theta}$ are given for a reduction of the  relative position, relative velocity and the pitch angle, respectively. The negative reward $r_{dur}$  gives the agent an incentive to start moving.
Considering the negative weights $w_{p},w_{v},w_{\theta}, w_{dur}$   and the agent frequency $f_{ag}$, derived in Sec.~\ref{sec:hyperparemter_estimation}, with which actions are drawn from $\mathbb{A}$, we define one time step as $\Delta t=1/f_{ag}$ and $r_p,r_v,r_{\theta},r_{dur}$ as 

\begin{small}
\begin{align}
&r_p =\text{clip}(w_p(|p_{x,t}|-|p_{x,t-1}|),-r_{p,max},r_{p,max}) \label{eq:clip_r_p}\\
&r_v = \text{clip}(w_v(|v_{x,t}|-|v_{x,t-1}|),-r_{v,max},r_{v,max})\label{eq:clip_r_v}\\
&r_ {\theta} = w_{\theta}( |\theta_{d,t}|-|\theta_{d,t-1}|)/\theta_{max} v_{lim}\label{eq:scale_r_theta}\\
&r_{dur} = w_{dur}  v_{lim}\Delta t.\label{eq:scale_r_dur}
\end{align}
\end{small}

The weights are  negative so that decreasing the relative position, relative velocity or the pitch angle yields positive reward values. Clipping  $r_p$ and $r_v$ to  $\pm r_{p,max}$ and $\pm r_{v,max}$, as well as scaling $r_{\theta}$ and $r_{dur}$ with $v_{lim}$ is necessary due to their influence on the value of $r_{max}$. $r_{max}$ denotes the maximum achievable reward in a non-terminal timestep if $v_x \leq v_{lim}$ and $a_x \leq a_{lim}$ and thus complies with the limits set by the  motion scenario described in Sect.~\ref{sec:curriculum_discretization} during the derivation of the sequential curriculum. To ensure the applicability of the curriculum also in situations in which that compliance is not given is why the clipping is applied in   \eqref{eq:clip_r_p} and   \eqref{eq:clip_r_v}. $r_{max}$  plays a role in the scaling of Q-values that is performed as part of the knowledge transfer during different curriculum steps. It is defined as follows

\begin{small}
\begin{align}
&r_{p,max} = |w_p| v_{lim} \Delta t,\\
& r_{v,max} = |w_v| a_{lim} \Delta t,\\
&r_{\theta,max} = |w_{\theta}| v_{lim} \Delta\theta/\theta_{max}, \\
&r_{dur,max} = w_{dur}  v_{lim}\Delta t,\\
&r_{max} = r_{p,max}+r_{v,max}+r_{\theta,max}+r_{dur,max}.
\end{align}
\end{small}

With $r_{max}$ derived and the weights $w_{suc}$ and $w_{fail}$, we can finally define the success and failure rewards as

\begin{small}
\begin{equation}
r_{suc} = w_{suc} r_{max}  ~~ \textrm{and}~~~  r_{fail} = w_{fail} r_{max}.\label{eq:r_suc_r_fail}
\end{equation}
\end{small}

All weights of the reward function were determined experimentally.

\subsubsection{Double Q-Learning}
We use the Double Q-Learning algorithm \citep{Hasselt2010} to address the problem of overestimating action values and  increasing training stabilty. Inspired by \citep{Even-Dar2003}, we apply a state-action pair specific learning rate that decreases the more often the state-action pair has been visited. 

\begin{small}
\begin{align}
\alpha(s_t,a_t) = \max\left\lbrace \left( n_c(s_t,a_t)+1\right)^{-\omega},\alpha_{min}\right\rbrace,
\end{align}
\end{small}

\noindent where $ n_c(s_t,a_t)$ denotes the number of visits the state action-pair $\left(s_t,a_t \right)$ has been experiencing until the timestep $t$ and $\omega$ is a decay factor. To keep a certain minimal learning ability, the learning rate is constant once the value of $\alpha_{min}$ has been reached. Sufficient exploration of the state space can be ensured by applying an $\epsilon$-greedy policy for the action selection during training. The exploration rate $\epsilon$ is  varied according to a schedule for the episode number.

\subsection{Curriculum and Discretization}\label{sec:curriculum_discretization}
Our curriculum and discretization approach builds on \citep{Lampton2009}, where a multiresolution method for state space discretization is presented. There, starting with a coarse discretization, a region around a pre-known learning goal is defined. Once the agent has learned to reach that region, it is further refined by discretizing it with additional states.  Then, training starts anew but only in the refined region, repeatedly until a pre-defined resolution around the learning goal has been reached. Consequently, a complex task is reduced to a series of smaller, quickly-learnable sub tasks.
\begin{figure*}
\centering
\includegraphics[scale = 0.45]{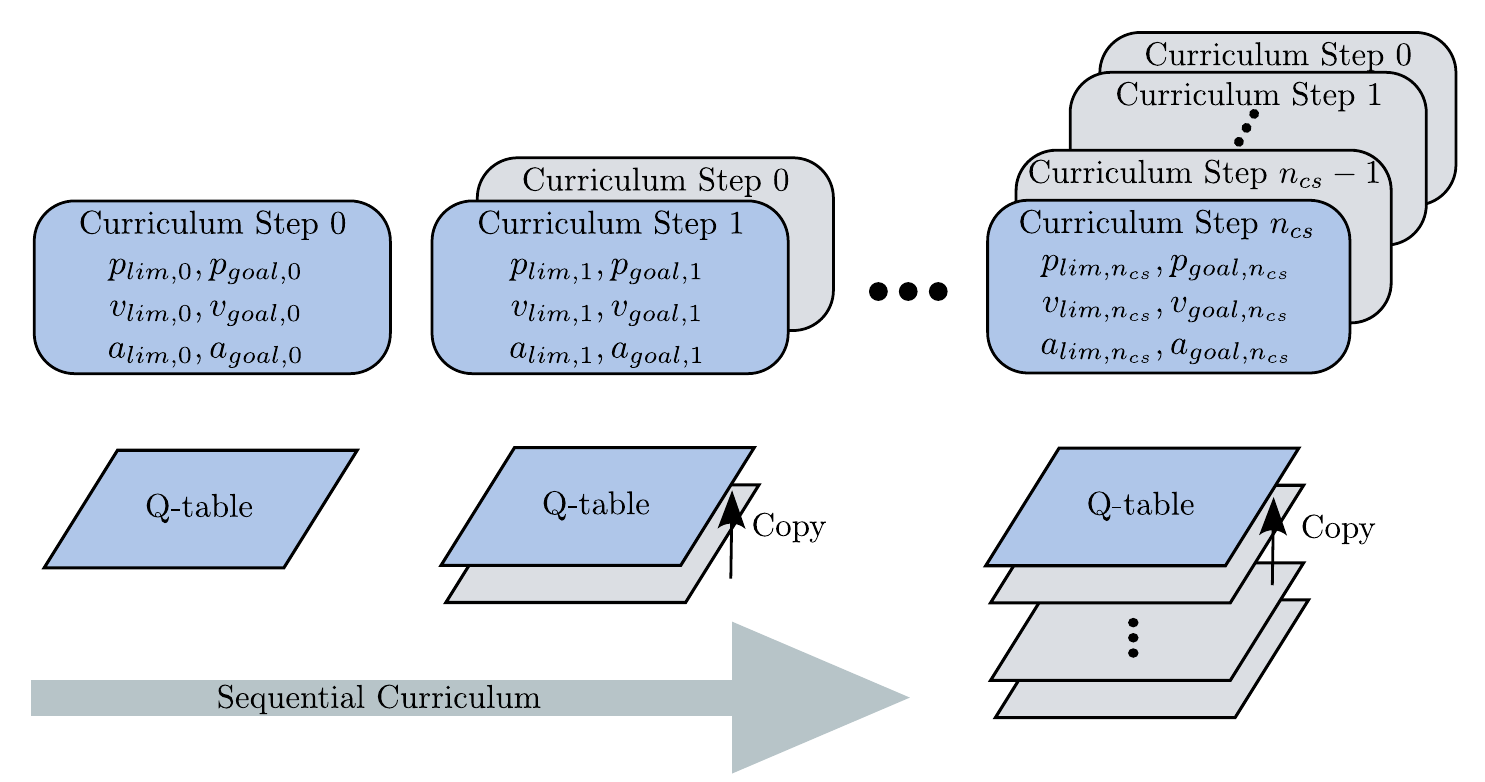}
\caption{Illustration of the sequential curriculum for $n_{cs}+1$ curriculum steps. Each curriculum step consists of an instance of the basis learning task.}
\label{fig:seq_cur}
\end{figure*}
In our novel approach, we introduce a similar multiresolution technique combined with transfer learning as part of a sequential curriculum \citep{Narvekar2020} to accelerate and stabilize training even further.
Each new round of training in the multiresolution setting constitutes a different curriculum step. A curriculum step is an instance of the basis learning task introduced in Sect.~\ref{sec:basis_learning_task}. This means in particular that each curriculum step has its own Q-table. Since the size of the state and action space is the same throughout all steps, knowledge can be easily transferred to the subsequent curriculum step by copying the Q-table that serves then as a starting point for its training. Fig.~\ref{fig:seq_cur} illustrates the procedure of the sequential curriculum.\\
However, throughout the curriculum the goal regions become smaller, leading to less spacious maneuvers of the multi-rotor vehicle. Thus, the achievable reward, which depends on relative change in position and velocity, is reduced. In order to achieve a consistent adaptability of the Q-values over all curriculum steps, the initial Q-values need to be scaled to match the maximum achievable reward per timestep. Therefore, we define

\begin{small}
\begin{equation}
Q_{init, i+1} = r_{max,i+1}/r_{max,i} Q_{result,i},
\end{equation}
\end{small}

\noindent where $i\geq 1$ is the current curriculum step. Furthermore, adding more curriculum steps with smaller goal regions possibly leads to previously unconsidered effects, such as overshooting the goal state. This is why in our work, training is not only restricted to the latest curriculum step but performed also on the states of the previous curriculum steps when they are visited by the agent. 

\begin{figure}
\centering
\includegraphics[scale = 0.3]{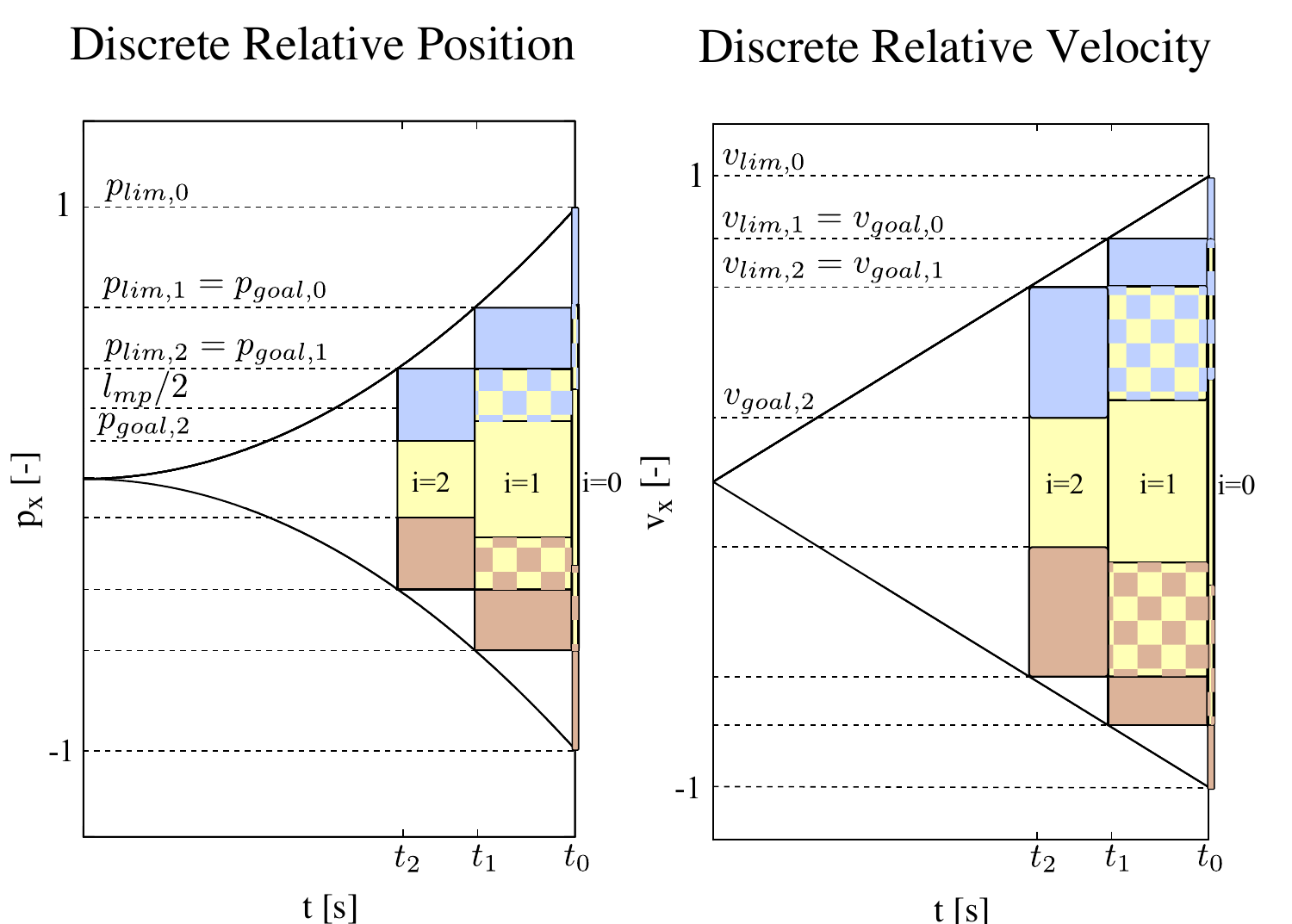}
\caption{Illustration of the mapping of the normalized observations $p_x$ and $v_x$ to the discrete states (red - $0$, yellow - $1$, blue - $2$) for a curriculum with three steps $(i=0,i=1,i=2)$. The yellow chequered regions illustrate the size of the goal state when the curriculum step is the last of the sequence.}
\label{fig:discretization}
\end{figure}

Furthermore, the discretization introduced in Sect.~\ref{sec:discrete_state_space},  which is now associated with a curriculum step, covers only a small part of the continuous observation space with few discrete states. In order to completely capture the predominant environment characteristics of the subspace in which the agent is expected to learn the task, it is necessary for it to have the knowledge of the state space topology \citep{Anderson1994}. Against the background  of the multiresolution approach, this can be rephrased as the question of how to define suitable goal regions, i.e., the size of the goal states. To this end, we again leverage the insight that accurate knowledge about the relative position and velocity is only required when landing is imminent. We begin by choosing an exponential contraction of the size of the unnormalized goal state $p_{c,x,goal,i}$ associated with the $i^{th}$ curriculum step such that 

\begin{small}
\begin{align}
&p_{c,x,goal,0} = \sigma^2 x_{max} ,\\
&p_{c,x,goal,i} = \sigma^{2(i+1)}x_{max},\\
&p_{c,x,goal,n_{cs}} = l_{mp}/2 = \sigma^{2(n_{cs}+1)}x_{max}, \label{eq:exp_contraction}
\end{align}
\end{small}

\noindent where $0<\sigma<1$ is a  contraction factor, $l_{mp} \in \mathbb{R}[\si{m}]$ is the edge length of the squared moving platform, $x_{max}$ is the boundary of the fly zone $\mathcal{F}$ (see Sect.~\ref{sec:prelim_notations}) and $n_{cs}$ is the number of curriculum steps following the initial step. The contraction factor $\sigma$ plays a role in how easy it is for the agent to reach the goal state. If it is set too low, the agent will receive success rewards only rarely, thus leading to an increased training time. \eqref{eq:exp_contraction} can be solved for $n_{cs}$ because $\sigma$, $l_{mp}$ and $x_{max}$ are known parameters. 
To determine the relationship between the contraction of the positional goal state  and the contraction of the velocity goal state, we need a kinematics model that relates these variables. For this purpose, we envision the following  worst case scenario for the relative motion \eqref{eq:non_linear_rel_motion}.
Assume that the vehicle hovers right above the platform which is located on the ground in the center of the fly zone $\mathcal{F}$, with $v_x = 0\si{m/s}$ and $\psi_{rel} = 0$. Now, the platform starts to constantly accelerate with the maximum possible acceleration $a_{mp,max}$ defined by \eqref{eq:a_mpmax} until it has reached $x_{max}$  after a time $t_{0} = \sqrt{2 x_{max}/a_{mp,max}}$.
This is considered to be the worst case since the platform never slows down, unlike the rectilinear periodic movement \eqref{eq:platform_acceleration_rpm} used for training.
The evolution of the continuous observations over time is then expressed by

\begin{small}
\begin{align}
&a_{x,c,wc}(t) = a_{mp,max},  \label{eq:a_p_s_rel_max}\\
& v_{x,c,wc}(t) =a_{mp,max} t, \\
& p_{x,c,wc}(t) =0.5 a_{mp,max}t^2,\label{eq:a_v_s_rel_max}
\end{align}
\end{small}

\noindent which constitute a simple kinematics model that relates the relative position and relative velocity via the time $t$. Next, we consider $p_{max}=x_{max}$, $v_{max}= a_{mp,max}t_{0}$ and $a_{max} = a_{mp,max}$ for the normalization \eqref{eq:xyz_c_x} - \eqref{eq:xyz_c_z} and discretize \eqref{eq:a_p_s_rel_max} - \eqref{eq:a_v_s_rel_max} via the time $t_i =  \sigma^i t_0$. This allows us  to determine the values of  $p_{lim,i}$, $v_{lim,i}$, $a_{lim,i}$  and $p_{goal,i}$, $v_{goal,i}$, $a_{goal,i}$. They  are required for the basis learning task presented in Sect.~\ref{sec:basis_learning_task} that now constitutes  the $i^{th}$ curriculum step. 
For this purpose, we define with $i \in \left\lbrace 0,1,\ldots,n_{cs}\right\rbrace$

\begin{small}
\begin{align}
p_{lim,i}&= \sigma^{2i}= 0.5 a_{mp,max} t_i^2/p_{max}, \\
v_{lim,i}&=\sigma^{i}= a_{mp,max} t_i /v_{max}, \\
a_{lim,i} &= 1 , \\
p_{goal,i}&=\beta_p \sigma^{2i}=\beta_p p_{lim,i}, \\
v_{goal,i}&=\beta_v\sigma^i=\beta_v v_{lim,i}, \\
a_{goal,i} &= \beta_{a}\sigma_{a}=\beta_{a}\sigma_{a} a_{lim,i} , 
\end{align}
\end{small}

\noindent where $\beta_p=\beta_v =\beta_{a} = 1/3$ if the $i$th curriculum step is the curriculum step  that was most recently added to the curriculum sequence, and $\beta_p=\sigma^2,\beta_v = \sigma,\beta_{a}=1$ otherwise.  Thus, for the latest curriculum step the goal values  result from scaling the associated limit value with a factor of $1/3$, a value that has been found empirically, and for all previous steps from the discretized time applied to \eqref{eq:a_p_s_rel_max} - \eqref{eq:a_v_s_rel_max}. The entire discretization procedure is illustrated by Fig.~\ref{fig:discretization}. 
Introducing a different scaling value only for the latest curriculum step has been empirically found to improve the agent's ability to follow the moving platform. The discretization of the acceleration is the same for all curriculum steps and is defined by a contraction factor $\sigma_a$ which  has been empirically set. Finding a suitable value for $\sigma_a$ has been driven by the notion that if this value is chosen too high, the agent will have difficulties reacting to changes in the relative acceleration. This is due to the fact that the goal state would cover an exuberant range in the discretization of the relative acceleration. However, if it is chosen too low the opposite is the case. The agent would only rarely be able to visit the discrete goal state $a_{goal,i}$, thus unnecessarily foregoing one of only three discrete states.  \\
During the training of the different curriculum steps the following episodic terminal criteria have been applied. On the one hand, the episode terminates with success \textcolor{black}{and the success reward $r_{suc}$ is received} if the goal state $s^*$ of the latest curriculum step is reached if the agent has been in that curriculum step's discrete states for at least one second without interruption. \textcolor{black}{This is different to all previous curriculum steps where the success reward $r_{suc}$ is received immediately after reaching the goal state of the respective curriculum step as is explained in Sect. \ref{sec:reward_function}.} On the other hand, the episode is terminated with a failure if the multi-rotor vehicle leaves the fly-zone $\mathcal{F}$ or if the success terminal criterion has not been met after the maximum episode duration $t_{max}$. The reward received by the agent in the terminal time step for successful or failing termination of the episode is defined by \eqref{eq:r_suc_r_fail}.
Once the trained agent is deployed, the agent selects the actions based on the Q-table of the latest curriculum step to which the continuous observations can be mapped (see Fig.~\ref{fig:seq_cur}).

\subsection{Hyperparameter Determination} \label{sec:hyperparemter_estimation}

We leverage the discrete action space \eqref{eq:discrete_action_space} to determine the hyperparameters  agent frequency $f_{ag}$ and maximum pitch angle $\theta_{max}$ in an interpretable way. The purpose is to ensure sufficient maneuverability of the UAV to enable it to follow the platform. For sufficient maneuverability, the UAV needs to possess  two core abilities. i) Produce a maximum acceleration bigger than the one the platform is capable of and ii) change direction of acceleration quicker than the moving platform. Against the background of the assumptions explained in Sect.~\ref{sec:basis_learning_task_preliminaries} complemented with thrust compensation,  we consider the first aspect by the maximum pitch angle $\theta_{max}$

\begin{small}
\begin{equation}
\theta_{max} = \tan^{-1}\left( \frac{k_a a_{mp,max}}{g}\right).\label{eq:theta_max}
\end{equation}
\end{small}
\begin{figure}[thpb]
      \centering
       \includegraphics[width=7cm]{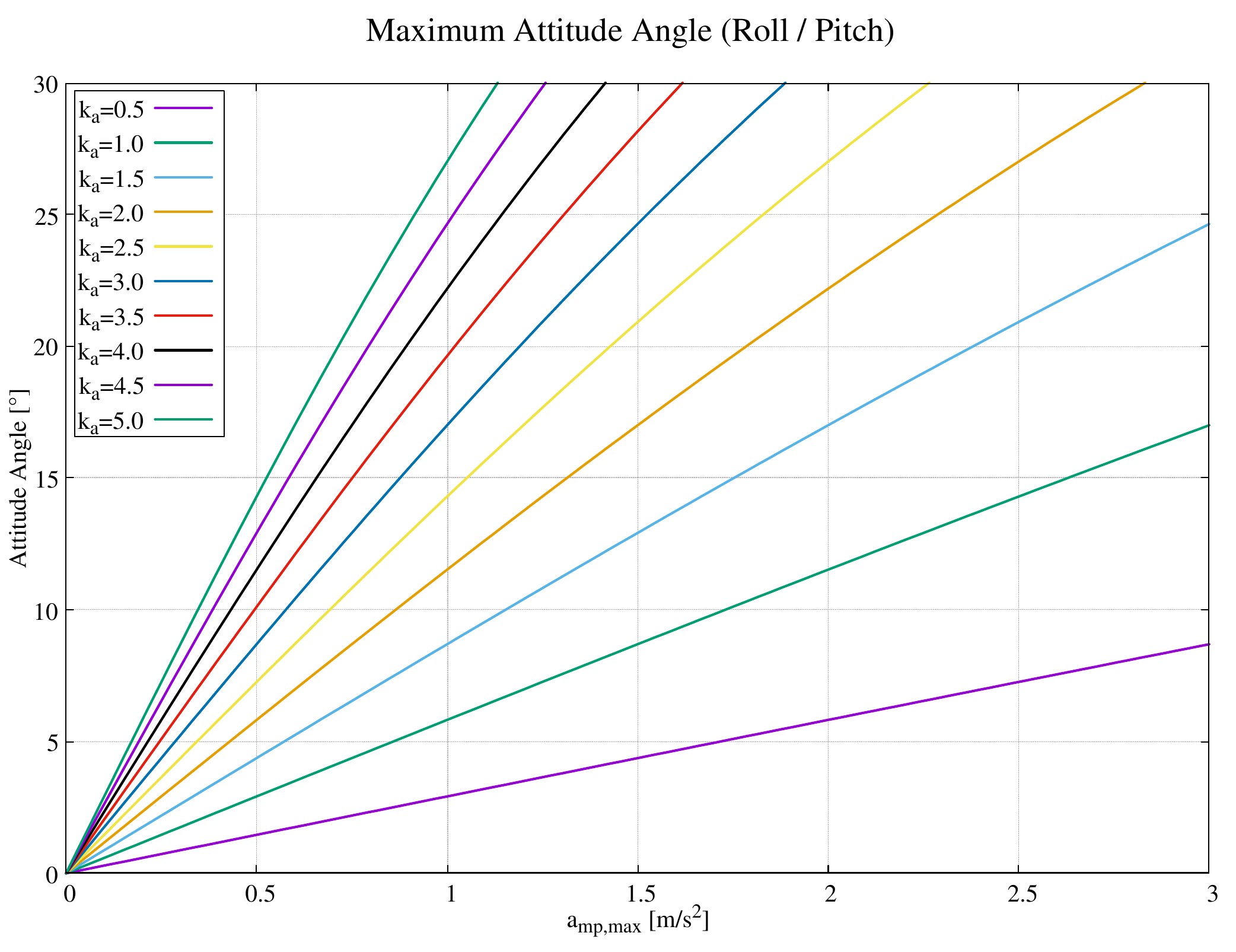}			
\caption{\textcolor{black}{Maximum pitch angle $\theta_{max}$ for different values of $k_a$ and $a_{mp,max}$.}}
\label{fig:max_attitude_angle}
 \end{figure}

$k_{a}$ denotes the multiple of the platform's maximum acceleration of which the UAV needs to be capable. \textcolor{black}{Fig. \ref{fig:max_attitude_angle} illustrates how different values for $k_a$ affect the maximum pitch angle $\theta_{max}$ for different values of the maximum platform acceleration $a_{mp,max}$}. For the second aspect, we leverage the platform's known frequency of the rectilinear periodic movement. According to \eqref{eq:platform_acceleration_rpm}, the moving platform requires the duration of one period to entirely traverse the available  range of acceleration. Leveraging equation \eqref{eq:normalized_pitch_set}, we can calculate the time required by the copter to do the same as $4n_{\theta}\Delta t$, where $\Delta t = 1/f_{ag}$. Next, we introduce a factor $k_{man}$ which specifies how many times faster the UAV should be able to roam through the entire range of acceleration than the moving platform. Against this background, we obtain the agent frequency as

\begin{small}
\begin{align}
f_{ag} = 4 n_{\theta}  k_{man}  \frac{\omega_{mp}}{2\pi}= 2n_{\theta}  k_{man} \frac{\omega_{mp}}{\pi}. \label{eq:f_ag}
\end{align}
\end{small}

\begin{figure}[thpb]
      \centering
       \includegraphics[width=7cm]{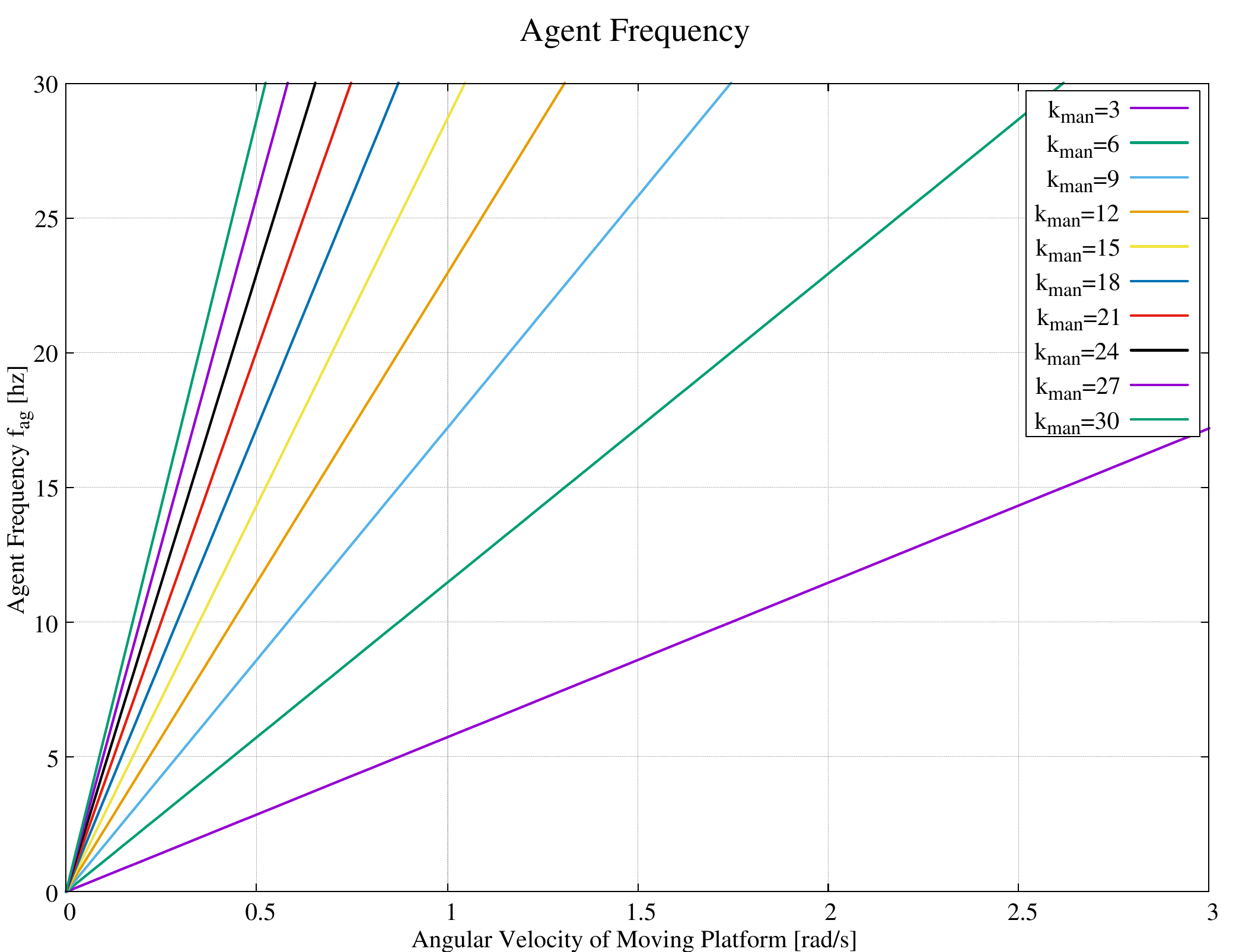}			
\caption{\textcolor{black}{Agent frequency $f_{ag}$  for different values of $k_{man}$ and $\omega_{mp,max}$.}}
\label{fig:agent_freq}
 \end{figure}
\textcolor{black}{Fig. \ref{fig:agent_freq} illustrates how the agent frequency $f_{ag}$ depends on the values of $k_{man}$ and the angular frequency of the platform $\omega_{mp}$ executing the rectilinear periodic movement.}\\
Both hyperparameters $\theta_{max}$ and $f_{ag}$ are eventually based on the maximum acceleration of the moving platform $a_{mp,max}$ as is the discretization of the state space. Therefore, we argue that these hyperparameters pose a matching set of values that is suitable to prevent excessive jittering in the agent's actions.

%% file: 04_implementation.tex
\section{Implementation}\label{sec:Implementation}

\subsection{General}
We set up the experiments to showcase the following.
\begin{itemize}
\item Empirically show that our method is able to outperform the \textcolor{black}{RL} baseline \citep{Rodriguez-Ramos2019} with regard to the rate of successful landings while  requiring a shorter training time.
\item Empirically show that our  method is able to perform successful landings for more complex platform trajectories such as an 8-shape.  
\item Demonstrate our method on real hardware.
\end{itemize}

\subsection{Simulation Environment} 
\begin{figure*}[thpb]
      \centering
		\includegraphics[scale=0.9]{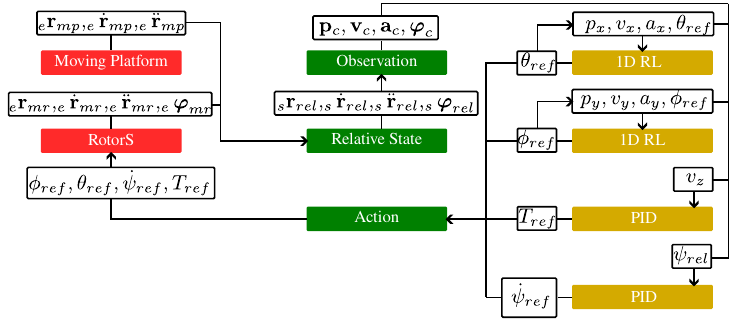}
\caption{Overview of the framework's structure. Red are components relying on Gazebo. Green are components designed for  converting and extracting / merging data. Yellow are components dealing with the control of the UAV. }
\label{fig:sim_framework}
 \end{figure*}

The  environment is built within the physics simulator Gazebo 11, in which the moving platform and the UAV  are realized. We use the RotorS simulator \citep{Furrer2016} to simulate the latter. Furthermore, all required tasks  such as computing the observations of the environment, the state space discretization as well as the Double Q-learning algorithm are implemented as nodes in the ROS Noetic framework using Python 3.\\
The setup and data flow are illustrated by Fig.~\ref{fig:sim_framework}. The associated parameters are given in Table~\ref{tab:environment_params}. We obtain $\mathbf{a}_c$ by taking the first order derivative of $\mathbf{v}_c$  and applying a first order Butterworth-Filter to it. The cut-off frequency is set to $0.3 \si{hz}$.

\begin{table*}[]
\centering
\fontsize{\tfs}{\bfs}\selectfont
\begin{tabular}{@{}ccccccccc@{}}
\toprule
Group & \multicolumn{2}{c}{Gazebo}                                                                                                 & \thead{Relative State} & \thead{Observations} & \thead{1D RL Agent } & \multicolumn{2}{c}{\thead{PID Controller }}                                                  \\ \midrule
Name  & \thead{Max. step \\size\\ $[\si{s}]$} & \thead{Real time \\ factor\\$[-]$} & \thead{{ }\\Freq. \\$[\si{hz}]$} & \thead{{ }\\Freq.\\ $[\si{hz}]$ } &\thead{Pitch / Roll\\ Freq. \\$[\si{hz}]$}  & \thead{Yaw\\Freq. $[\si{hz}]$ \\ $k_p,k_i,k_d$} & \thead{$v_z$\\Freq. $[\si{hz}]$\\ $k_p,k_i,k_d$} \\ \midrule
Value &$ 0.002$& $1$& $100$  & $100$ &\thead{see\\ \eqref{eq:f_ag}}                                                    &\thead{$\sim 110$\\ $8,1,0$} & \thead{$\sim 110$\\ $5,10,0$} \\ \bottomrule
\end{tabular}
\caption{Parameters of the training environment.}
\label{tab:environment_params}
\end{table*}

\subsection{Initialization}
In \citep{Kooi2021}, it is indicated that  training can be accelerated when the agent is initialized close to the goal state at the beginning of each episode. For this reason, we use the following normal distribution to determine the UAV's initial position within the fly zone $\mathcal{F}$ during the first curriculum step.

\begin{small}
\begin{equation}
(x_{init},y_{init}) = \left(\text{clip}\left(N(\mu,\sigma_{\mathcal{F}}),-x_{max},x_{max},\right),0\right)
\end{equation}
\end{small}
We set $\sigma_{\mathcal{F}} = p_{max}/3$, which will ensure that the UAV is initialized close to the center of the flyzone  and thus in proximity to the moving platform more frequently. All subsequent curriculum steps as well as the testing of a fully trained agent are then conducted using a uniform distribution over the entire fly zone.

\subsection{Training Hardware}
Each training is run on a desktop computer with the following specifications. Ubuntu 20, AMD  Ryzen threadripper 3960x 24-core processor, 128 GB RAM, 6TB SSD. This allows us to run up to four individual trainings in parallel. Note, that being a tabular method, the Double Q-learning algorithm does not depend on a powerful GPU for training since it does not use a neural network.

\subsection{Training} \label{sec:training}
We design two training cases, \textit{simulation} and \textit{hardware}. Case \textit{simulation} is similar  to the training conditions in the baseline method so that we can compare our approach in simulation. Case \textit{hardware} is created to match the spatial limitations of our real flying environment, so that we can evaluate our approach on real hardware. 
For both cases, we apply the rectilinear periodic movement (RPM) of the platform specified by \eqref{eq:platform_acceleration_rpm} during training. We consider different scenarios for training regarding the maximum velocity of the platform, which are denoted by ``RPM $v_{mp,train}$''.  For each velocity $v_{mp,train}$, we train  four agents using the same parameters.  The purpose is to provide evidence of reproducibility of the training results instead of only presenting a manually selected, best result. We choose the same UAV (Hummingbird) of the RotorS package that is also used in the baseline method. Other notable differences between the baseline and the training  cases \textit{simulation} and \textit{hardware} are summarized in Table~\ref{tab:training_differences}. 

\begin{table*}
\fontsize{\tfs}{\bfs}\selectfont
\centering
\begin{tabular}{ccccccc}
\toprule
Method & \thead{Fly zone size $[m]$} & \thead{Platform size $[m]$} & \makecell{$r_{mp} [\si{m}]$} & \thead{$v_{mp} [\si{m/s}]$} & \thead{$a_{mp,max} [\si{m/s^2}]$} & \thead{$f_{ag} [hz]$}\\ \hline
Baseline & $3\times 6$ & $1\times1\times\sim 0.3$ & $\sim 2.5$ & $1$  &$\sim 0.4$ & $20$  \\ \hline
\multirow{3}{*}{Case \textit{simulation}} &\multirow{3}{*}{$9\times9$}&\multirow{3}{*}{$1\times1\times0.3$}& \multirow{3}{*}{2} &  $0.8$ & $0.32$ & $11.46$ \\
 									& & &  & $1.2$ & $0.72$ & $17.19$ \\
 									& & &  & $1.6$ & $1.28$ & $22.92$ \\\hline
Case \textit{hardware}&$2\times2$ &$0.5\times0.5\times0.3$& $0.5$& $0.4$& $0.32$ & $22.92$\\
\bottomrule
\end{tabular} 
\caption{Differences between the training parameters of our method and the baseline.}
\label{tab:training_differences}
\end{table*}

Note, that some of our training cases  deal with a higher maximum  acceleration of the moving platform  than the training case used for the baseline method and are therefore considered as more challenging. For all trainings, we use an initial altitude for the UAV of $z_{init} = 4\si{m}$ and a vertical velocity of $v_z = -0.1\si{m/s}$ so that the UAV is descending during an entire episode. The values used for these variables in the baseline method can not be inferred from the paper. For the first curriculum step, the exploration rate schedule is empirically set to  $\varepsilon= 1$ (episode $0 - 800$) before it is linearly reduced to $\varepsilon=0.01$ (episode $800-2000$). For all later curriculum steps, it is $\varepsilon=0$. For choosing these values, we could exploit the discrete state-action space where for each state-action pair the number of visits of the agent can be tracked. The exploration rate schedule presented above is chosen so that most state action pairs have at least received one visit.
Other training parameters are presented in Table~\ref{tab:common_training_params}. 
The training is ended as soon as the agent manages to reach the goal state \eqref{eq:goal_state} associated with the latest step in the sequential curriculum in $96\%$ of the last $100$ episodes. This value has been empirically set. For all trainings, we use noiseless data to compute the observations fed into the agent. However, we test selected agents for their robustness against noisy observations as part of the evaluation. 

\begin{table*}[]
\fontsize{\tfs}{\bfs}\selectfont
\centering
\begin{tabular}{@{}ccccccccc@{}}
\toprule
Group & \multicolumn{2}{c}{\begin{tabular}[c]{@{}c@{}}Double  Q-Learning\end{tabular}}                                               & \multicolumn{3}{c}{Reward function}           & \multicolumn{3}{c}{\begin{tabular}[c]{@{}c@{}}Discretization\end{tabular}} \\ \midrule
Name  & \begin{tabular}[c]{@{}c@{}}$\gamma$\end{tabular} & \begin{tabular}[c]{@{}c@{}}$\alpha_{min}$ \\$\omega$\end{tabular} &\begin{tabular}[c]{@{}c@{}}$w_p$\\ $w_v$ \\ $w_{\theta}$\end{tabular}   & \begin{tabular}[c]{@{}c@{}}$w_{dur}$\\ $w_{suc}$ \\ $w_{fail}$\end{tabular} &$t_{max}$& \begin{tabular}[c]{@{}c@{}}$\sigma_{a}$\\$\sigma $\end{tabular}   &  \begin{tabular}[c]{@{}c@{}}$n_{\theta}$\\$n_{cs,sim.}$\\$n_{cs,hardw.}$\end{tabular} & \begin{tabular}[c]{@{}c@{}}$k_{a}$\\$k_{man}$\end{tabular}    \\ \midrule
Value & $0.99$                                                      &  \begin{tabular}[c]{@{}c@{}} $0.02949$ \\$0.51$\end{tabular} &\begin{tabular}[c]{@{}c@{}}$-100$\\ $-10$ \\ $-1.55$\end{tabular}  & \begin{tabular}[c]{@{}c@{}}$-6$\\$ 2.6$\\ $-2.6$\end{tabular}    &$20\si{s}$ &\begin{tabular}[c]{@{}c@{}}$0.416$\\$0.8 $\end{tabular}             &   \begin{tabular}[c]{@{}c@{}}$3$\\$4$\\$3$\end{tabular}   &  \begin{tabular}[c]{@{}c@{}}$3$\\$15$\end{tabular}         \\ \bottomrule
\end{tabular} 
\caption{Training parameters of our approach that are associated with the Double Q-Learning algorithm, the reward function or the discretization of the state space.}
\label{tab:common_training_params}
\end{table*}

\subsection{Initiation of Motion} \label{sec:state_lock}
\begin{figure*}
\centering
\includegraphics[scale = 0.5]{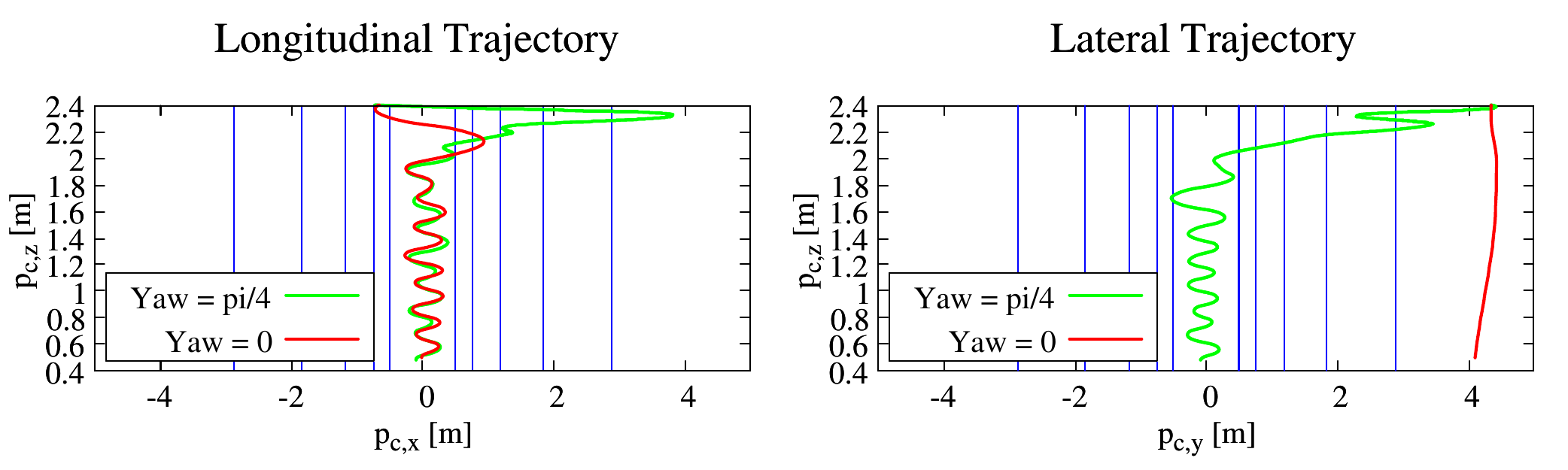}
\caption{Illustration of the problem of motion initiation. After initialization, commanding a yaw angle of $\psi_{rel} = \pi/4$ allows the lateral agent to enter a state associated with another action than "do nothing" due to the state change induced by the longitudinal platform movement. The platform movement is now reflected in $p_y,v_y,a_y$ that are the observations fed into the lateral agent.}
\label{fig:state_lock}
\end{figure*}

During training, the yaw controller ensures $\psi_{rel} = 0$. However, during evaluation in simulation this sometimes led to the situation that the agent commanding the lateral motion of the UAV  occasionally was not able to leave its initial state (see Fig.~\ref{fig:state_lock}). We hypothesize that the reason for this behavior is that the agent is trained on a platform that \textit{moves} while considering  \textit{relative} motion in its observations.  As a consequence, the agent learns a policy that, while being in certain states, exclusively relies on the  platform movement to achieve a desired state change. However, for evaluation using  $\psi_{rel} = 0$, the agent controlling the lateral motion observes no relative movement if the UAV is hovering and the platform following a rectilinear trajectory in longitudinal direction.
The issue can be addressed by setting initial $\psi_{rel} \neq 0$. 
In this case,  the  platform's movement shows a component in the values  ${p}_{c,y},{v}_{c,y}$ and ${a}_{c,y}$ that are used as observations for the lateral motion's agent. A change in the states is therefore much more likely, allowing the agent to enter states in which the policy selects an action other than ``do nothing''.  For this reason, we apply $\psi_{rel} = \pi/4$ for all experiments.

\textcolor{black}{\subsection{Cascaded PI Controller Baseline}
In order to be able to compare the performance of our approach with a non-learning based control method, we implemented a cascaded PI controller with two cascades as is illustrated by Fig. \ref{fig:cascaded_pid_structure}.
 \begin{figure*}[thpb]
\centering
\includegraphics[width=12cm]{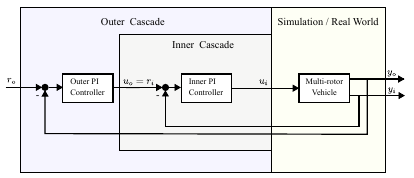}
\caption{\textcolor{black}{Structure of the cascaded PI controller. }}
\label{fig:cascaded_pid_structure}
 \end{figure*}
 We apply separate instances for controlling the longitudinal, lateral and vertical movement of the UAV. The properties of the different instances are summarized in Table~\ref{tab:props_cascaded_pid_controller}. The outer cascade tracks the setpoint $0\si{m}$ for the relative position $_sr_{rel}$ between the multi-rotor vehicle and the moving platform. For this purpose, it generates a setpoint for the relative velocity $_s\dot{r}_{rel}$ that is tracked by the inner cascade. The inner cascade aims to reach the velocity setpoint by computing setpoints for the attitude angle $\phi_{ref},\theta_{ref}$ or the thrust $T_{ref}$, respectively. The attitude angle or thrust is then tracked by the multi-rotor vehicle’s low level controllers. In order to achieve comparability between our RL approach and the cascaded PI controller, we limit the velocity setpoints generated by the outer cascade to the value of $v_{max} = a_{mp,max}t_0$ (see Sect. \ref{sec:curriculum_discretization}) and the attitude angles generated by the inner cascade to the same angles determined using \eqref{eq:theta_max} for the training scenario associated with the agent chosen for comparison.
\begin{table*}[]
\fontsize{\tfs}{\bfs}\selectfont
\centering
\begin{tabular}{|c|c|c|c|}
\Xhline{1\arrayrulewidth}
Controller & Cascade property  & Outer cascade (index $o$)& Inner cascade (index $i$) \\
\Xhline{1\arrayrulewidth}
\multirow{6}{*}{Longitudinal}&\fontsize{\tfs}{\bfs}\selectfont Controlled variable $y_o,y_i$   & $_sr_{rel,x}$ &$_s\dot{r}_{rel,x}$ \\
&\fontsize{\tfs}{\bfs}\selectfont Setpoint  $r_o,r_i$   & $0\si{m}$ &$_s\dot{r}_{rel,x,ref}$ \\ 
&\fontsize{\tfs}{\bfs}\selectfont Control effort $u_o,u_i$  &$_s\dot{r}_{rel,x,ref}$ & $\theta_{ref}$ \\ 
&\fontsize{\tfs}{\bfs}\selectfont  Limits of control effort $u_o,u_i$  & $-v_{max},v_{max}$ & $-\theta_{max},\theta_{max}$ \\ 
&\fontsize{\tfs}{\bfs}\selectfont Gains $k_p,k_i,k_d$ & $3,0,0$ & $8,1,0$ \\ 
& Wind up limit   & $5$ & $5$ \\ 
\Xhline{1\arrayrulewidth}
\multirow{6}{*}{Lateral}&\fontsize{\tfs}{\bfs}\selectfont Controlled variable $y_o,y_i$   & $_sr_{rel,y}$ &$_s\dot{r}_{rel,y}$ \\
&\fontsize{\tfs}{\bfs}\selectfont Setpoint $r_o,r_i$   &  $0\si{m}$ &$_s\dot{r}_{rel,y,ref}$ \\ 
&\fontsize{\tfs}{\bfs}\selectfont Control effort $u_o,u_i$  &$_s\dot{r}_{rel,y,ref}$ & $\phi_{ref}$ \\ 
&\fontsize{\tfs}{\bfs}\selectfont  Limits of control effort $u_o,u_i$  & $-v_{max},v_{max}$ & $-\phi_{max},\phi_{max}$ \\ 
&\fontsize{\tfs}{\bfs}\selectfont Gains $k_p,k_i,k_d$ & $3,0,0$ & $-8,-1,0$ \\ 
& Wind up limit   & $5$ & $5$ \\ 
\Xhline{1\arrayrulewidth}
\multirow{6}{*}{Vertical}&\fontsize{\tfs}{\bfs}\selectfont Controlled variable $y_o,y_i$   & $_sr_{rel,z}$ &$_s\dot{r}_{rel,z}$ \\
&\fontsize{\tfs}{\bfs}\selectfont Setpoint $r_o,r_i$    & $0\si{m}$ &$_s\dot{r}_{rel,z,ref}$\\ 
&\fontsize{\tfs}{\bfs}\selectfont Control effort $u_o,u_i$  &$_s\dot{r}_{rel,z,ref}$ & $T_{ref}$ \\ 
&\fontsize{\tfs}{\bfs}\selectfont  Limits of control effort $u_o,u_i$  & $-0.1m/s,0.1m/s$ & $0N,20N$ \\ 
&\fontsize{\tfs}{\bfs}\selectfont Gains $k_p,k_i,k_d$ & $1,0,0$ & $-4,-2,0$ \\ 
& Wind up limit   & $5$ & $5$ \\ 
\Xhline{1\arrayrulewidth}
\end{tabular} 
\caption{\textcolor{black}{Properties of the cascaded PI controllers.}}
\label{tab:props_cascaded_pid_controller}
\end{table*} 
}

%% file: 05_experiments.tex
\section{Results}\label{sec:Experiments} 

\subsection{Evaluation in Simulation without Noise}
In this  scenario, all agents trained for case \textit{simulation} and \textit{hardware} are evaluated as well in simulation using noiseless observations of the environment, just as in training. Besides a static platform, we use two types of platform trajectories. The first is the rectilinear periodic movement (RPM) specified by \eqref{eq:platform_acceleration_rpm} and the second is an eight-shaped trajectory defined by 

\begin{small}
\begin{align}
    _e\mathbf{r}_{mp} &=r_{mp}\left[\sin(\omega_{mp} t),\sin(0.5\omega_{mp} t),0\right]^T, \\
     \omega_{mp} &= v_{mp}/r_{mp}.
\label{eq:platform_acceleration}
\end{align}
\end{small}

For all landing attempts, we specify an initial altitude of $z_{init} = 2.5\si{m}$. {A landing attempt is ended once the UAV touches the surface of the moving platform or reaches an altitude that is lower than the platform surface, i. e. misses the platform. If the center of the UAV is located above the moving platform at the moment of touchdown, the landing trial is considered successful. The value of  $z_{init}$ leads to a  duration of a landing attempt which corresponds roughly to the time $t_{max}$ used as maximum episode length during training, see reward function \eqref{eq:reward}. The information regarding the training duration of the agents is summarized in Tab.~\ref{tab:duration_results} for case \textit{simulation} and case \textit{hardware}. \\
The training durations of the different curriculum steps suggest that the majority of the required knowledge is learned during the first curriculum step. The later curriculum steps required significantly fewer episodes to reach the end condition. This is because the exploration rate is $\varepsilon=0$. Thus, the agent is only exploiting previously acquired knowledge, which is also supported by the accumulated sum of rewards (Fig.~\ref{fig:rewards}). This also implies that the decomposition of the landing procedure into several, similar sub tasks is a suitable approach to solve the problem.
The achieved success rates in simulation with noiseless observations of the environment are presented 
in Fig.~\ref{fig:success_rates} for training case \textit{simulation} and in Table~\ref{tab:success_hardware_case} for training case \textit{hardware}.
 \aboverulesep = 0.0mm
  \belowrulesep = 0.0mm
\begin{table*}[]
\fontsize{\tfs}{\bfs}\selectfont
\centering
\begin{tabular}{|c|c|c|c|c|c|}
\hline
\thead{Case $\rightarrow$}   & \multicolumn{3}{c|}{Case \textit{simulation}} & Case \textit{hardware} \\ 
	\cmidrule{2-5}
\thead{Curricul.\\ step$\downarrow$}   & RPM 0.8 &  RPM 1.2 &  RPM 1.6 &  RPM 0.4 \\ \hline 
0	                    & $112,109,112,119$ & $102,87,87,91$  & $92,83,88,90$	& $ 89,92,107,110$\\
						\hline 
1	                    & $7,6,7,6$	         & $6,5,5,5$       & $5,4,6,6$       & $5,4,5,6$ \\
						\hline 
2	                    & $7,6,7,6$	         & $7,5,6,5$       & $6,5,9,6 $	  & $5,6,5,7 $	 \\
						\hline 	
3	                    & $9,7,8,8$	         & $8,6,7,7$       & $7,6,8,8$	      &  $6,7,6,8$ \\
						\hline 								 			
4	                    & $8,8,10,9$        & $10,8,8,9$      & $8,8,8,9$	    &N.A.	 \\
						\hline 								
\multirow{2}{*}{Total}  &$143,136$	         & $133,111$	    & $118,106$     &$105,109$ \\
						 &$144,148$	         & $113,117$	    & $119,119$     &$124,133$ \\
						\hline  						
\end{tabular} 
\caption{Training time in minutes required for the different curriculum steps by four agents trained with identical parameters for the different training cases. The values are rounded to the nearest integer. The training scenarios are identified by the platform's rectilinear periodic movement with a maximum velocity of $v_{mp,train} = 0.8$ (RPM 0.8), $v_{mp,train} = 1.2$ (RPM 1.2) and $v_{mp,train} = 1.6$ (RPM 1.6) for case \textit{simulation} and $v_{mp,train} = 0.4$ (RPM 0.4) for  case \textit{hardware}.}
\label{tab:duration_results}
\end{table*} 
\begin{figure*}[thpb]
\centering
\includegraphics[width=10cm]{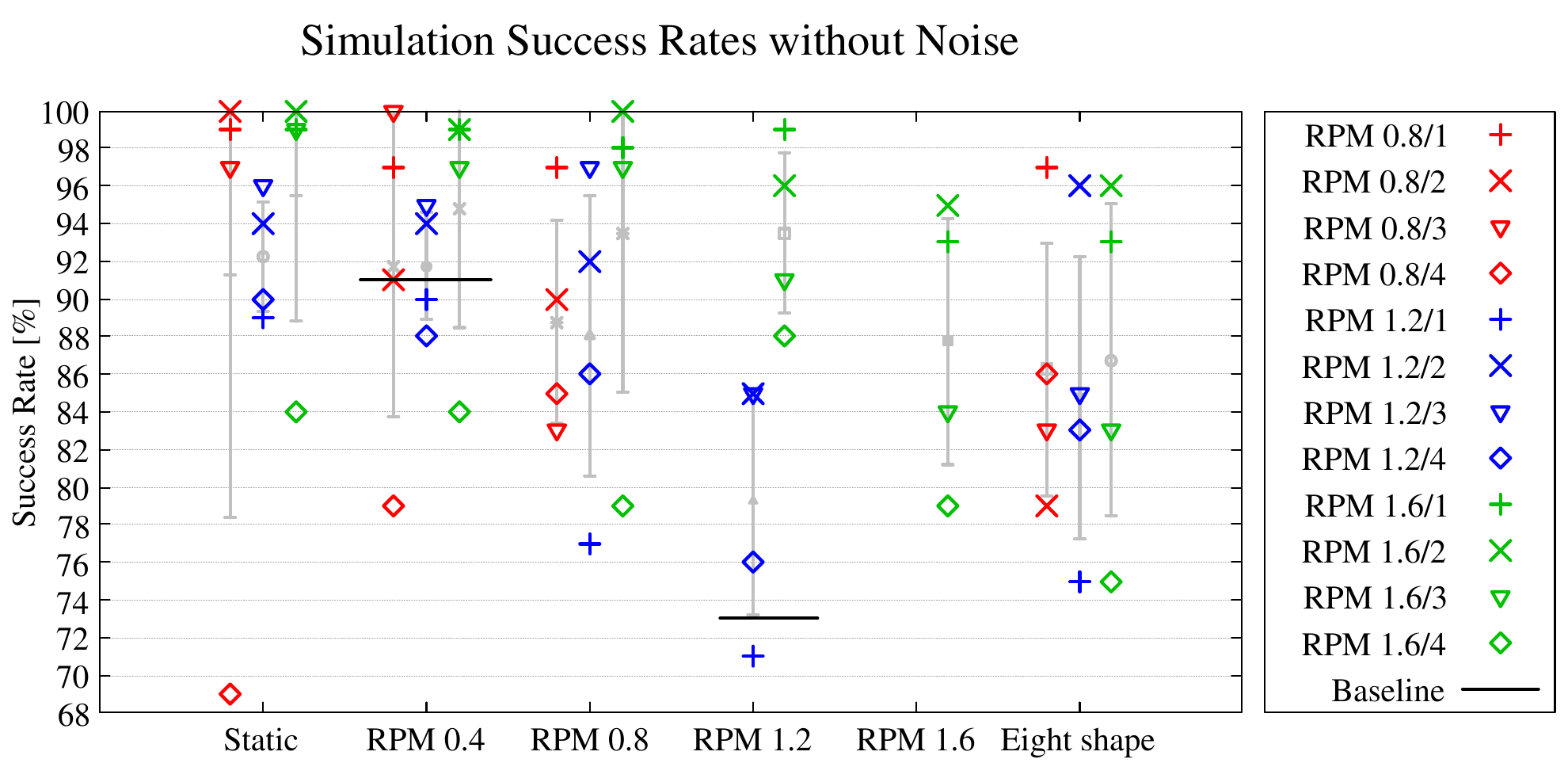}
\caption{Success rates of four agents trained for the case \textit{simulation} with  same parameters on a platform's rectilinear periodic movement with $v_{mp,train} = 0.8\si{m/s}$ (RPM 0.8 - red), $v_{mp,train} = 1.2\si{m/s}$ (RPM 1.2 - blue) and $v_{mp,train} = 1.6\si{m/s}$ (RPM 1.6 - green). The agents are evaluated on different types of platform movement, indicated by the values of the abscissa. Error bars illustrating the mean success rate and standard deviation are depicted in grey. The success rates have been determined for 150 landing trials for each evaluation scenario. }
\label{fig:success_rates}
 \end{figure*}
 \begin{table*}[thpb]
\fontsize{\tfs}{\bfs}\selectfont
\centering
\begin{tabular}{|c|c|c|c|c|c|c|}
\hline
Sim. eval. $\rightarrow$   & Static $[\%]$ &  RPM 0.4 $[\%]$ & RPM  0.8 $[\%]$ & RPM  1.2 $[\%]$ & RPM  1.6 $[\%]$  &8-shape $[\%]$ \\ 
\hline
\thead{Cascaded PI \\controller}	& $100$ & $100$ & $100$	& $ 100$	& $ 100$	& $ 100$\\
\hline   	   						
\end{tabular} 
\caption{\textcolor{black}{Success rates in percent achieved in simulation with noiseless observations by   the cascaded PI controller. It has been evaluated for six different scenarios of platform movement, as is indicated by the column titles.} }
\label{tab:success_cascaded_pid_noiseless_case_sim}
\end{table*} 
They were determined over a larger set of RPMs than in the baseline method and indicate a good performance of the approach. The success rates become higher when the platform velocity is lower during evaluation compared to the one applied during training. 
This is to be expected since the equations presented in Sec.~\ref{sec:hyperparemter_estimation} for hyperparameter determination ensure a sufficient maneuverability up to the velocity of the  rectilinear periodic movement used in training. 
For the training case RPM 0.8, the fourth agent has a comparably poor performance. For the static platform the agent  occasionally suffers from the problem of motion initiation as described in Sect.~\ref{sec:state_lock}, despite setting $\psi_{rel} = \pi/4 \si{rad}$. For the evaluation case where the platform moves with RPM 0.4, the reason is an oscillating movement, which causes the agent to occasionally overshoot the platform. A similar problem arises for agent four of training case \textit{hardware} where the agent occasionally expects a maneuver of the platform during the landing procedure. As a consequence, this agent achieves a success rate of only $82\%$ for a static platform. However, for higher platform velocities, the success rate improves significantly. \\

\subsection{Selection of Agents for further Evaluation}\label{sec:agent_selection}
We select the first agent of training case \textit{simulation} which is performing best over  all evaluation scenarios and was trained on a rectilinear periodic movement with a platform velocity of $v_{mp}=1.6\si{m/s}$. It is denoted  RPM 1.6/1. Its reward curve is depicted in Fig.~\ref{fig:rewards}. 
\begin{figure*}[thpb]
\centering
\includegraphics[scale=0.28]{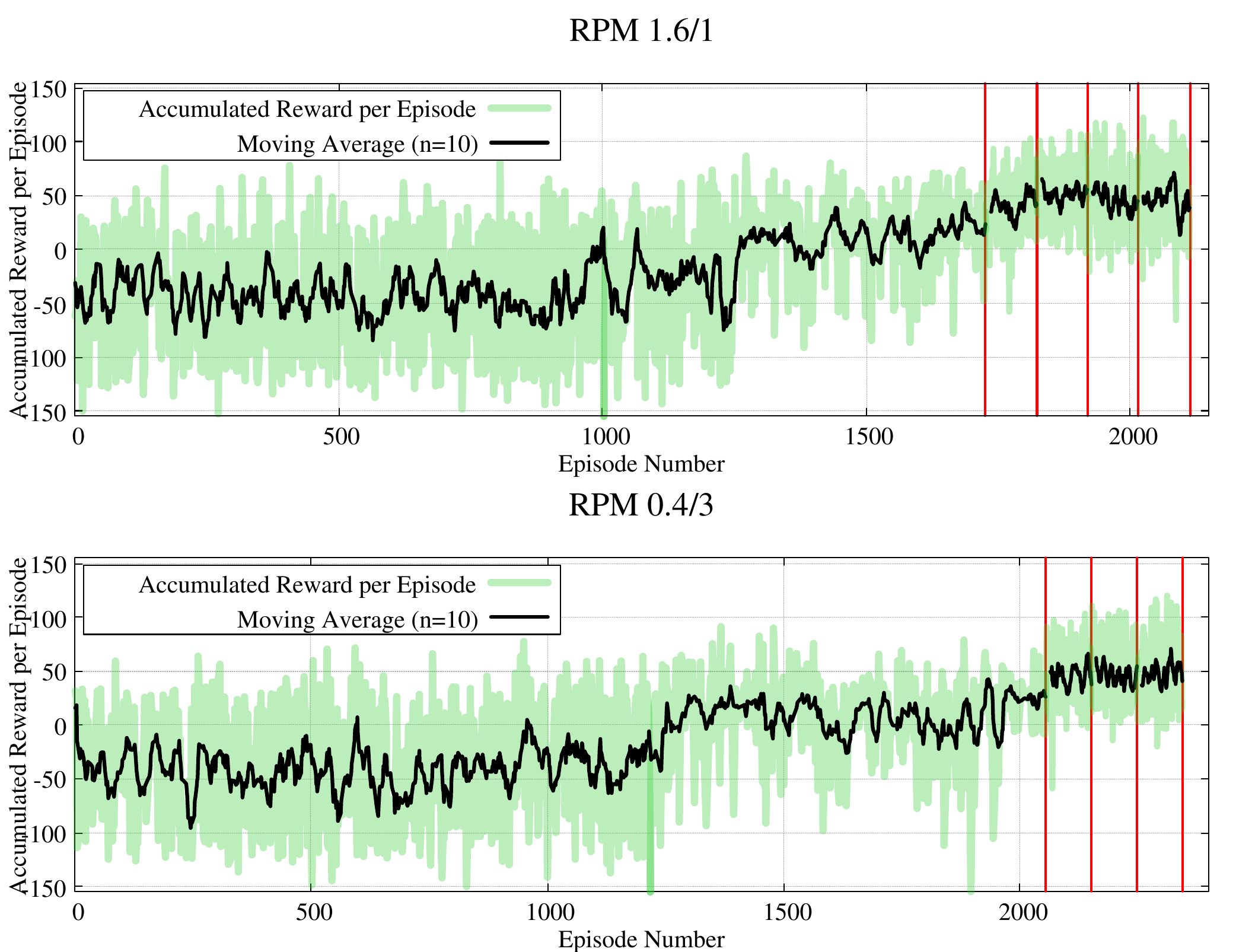}
\caption{Accumulated reward achieved  by agent RPM 1.6/1  of   case \textit{simulation}  (top) and agent  RPM 0.4/3 of  case \textit{hardware} (bottom) during training. Red lines indicate the end of a curriculum step. }
\label{fig:rewards}
 \end{figure*}
We  compare its results with the \textcolor{black}{RL} baseline method \textcolor{black}{and the cascaded PI controller} in Table~\ref{tab:comparison}. The comparison shows that our approach is able to outperform the \textcolor{black}{RL}  baseline method. For the RPM $0.4$ evaluation scenario with noiseless observations we achieve a success rate of $99\%$, which is $+8\% $ better than the baseline. For the RPM $1.2$ evaluation scenario, our method is successful in $99\%$ of the landing trials, increasing the \textcolor{black}{RL} baseline's success rate by $+26\%$.  Note that  the maximum radius of the rectilinear periodic movement is $r_{mp} = \sim2.5\si{m}$ in the \textcolor{black}{RL} baseline and $r_{mp}=2\si{m}$ in our approach. However, the value used for our approach poses a more difficult challenge since the acceleration acting on the platform is higher, due to the same maximum platform velocity, see Table~\ref{tab:training_differences}. Furthermore, our method requires $\sim 80\%$ less time to train and $53\%$ less episodes. \textcolor{black}{Comparison with the cascaded PI controller shows that our method is capable of similar performance than the cascaded PI controller in terms of success rates achieved in the two comparison scenarios. In order to achieve comparability in the first place, we limited the control efforts of the longitudinal and lateral controller's outer and inner cascade (see Table \ref{tab:props_cascaded_pid_controller}) to values also derived for the training case RPM 1.6. We obtain $u_o =v_{max} = a_{mp,max}t_0=3.39\si{m/s}$ as described in Sect. \ref{sec:curriculum_discretization} and $u_i = \theta_{ref} = \phi_{ref} = 21.38^\circ$ by means of \eqref{eq:theta_max}, respectively.  As is further shown by Table~\ref{tab:success_cascaded_pid_noiseless_case_sim} the cascaded PI controllers achieve a success rate of $100\%$ for all types of platform movement. However, the advantage of our learning based method is that it does not require a tuning procedure as is often conducted manually for cascaded PI controllers and therefore time consuming. This is compensated by a smoother flight behaviour enabled by the use of a continuous value range for the control effort of the PI controller. }\\ \textcolor{black}{For training case \textit{hardware} we select agent 3 for further evaluation, denoted RPM 0.4/3}. Its reward curve is also depicted in Fig.~\ref{fig:rewards}. It is able to achieve a success rate of $99\%$ for the evaluation scenario with a static platform, $100\%$ in case of a platform movement of RPM $0.2$,  $99\%$ for RPM $0.4$ and $97\%$ for the eight-shaped trajectory of the platform. It required  $2343$ episodes to train which took $123\si{min}$. \textcolor{black}{As already in case \textit{simulation}, the cascaded PI controller achieves a comparable performance across all types of platform movement as is shown by Table \ref{tab:success_hardware_case}. The values for the limits of the control efforts of the longitudinal and lateral controller's outer and inner cascade have been calculated as  $u_o =v_{max} = a_{mp,max}t_0=0.8\si{m/s}$ and $u_i = \theta_{ref} = \phi_{ref} = 5.59^\circ$}
\begin{table*}[thpb]
\fontsize{\tfs}{\bfs}\selectfont
\centering
\begin{tabular}{|c|c|c|c|c|c|}
\hline
Sim. eval. $\rightarrow$   & Static $[\%]$ &  RPM 0.2 $[\%]$ & RPM  0.4 $[\%]$ &  8-shape $[\%]$ \\ 
\hline
\thead{Case \\ \textit{hardware}}	& $100,100,99,82$ & $99,100,100,90$ & $98,99,99,97$	& $ 99,96,97,94$\\
\hline
\thead{Cascaded PI \\controller}	& $100$ & $100$ & $99$	& $ 100$\\
\hline   	   						
\end{tabular} 
\caption{\textcolor{black}{Success rates in percent achieved in simulation with noiseless observations by  four agents trained with same parameters for case \textit{hardware} and the cascaded PI controller. They have been evaluated for four different scenarios of platform movement, as is indicated by the column titles.} }
\label{tab:success_hardware_case}
\end{table*} 
 
\begin{table*}[]
\fontsize{\tfs}{\bfs}\selectfont
\centering
\begin{tabular}{|c|c|c|c|c|c|}
\hline
\thead{Category$\rightarrow$\\ Method$\downarrow$ }  & \thead{Fly zone \\size} &  \thead{Training\\ duration} & \thead{Success rate\\RPM 0.4}  &  \thead{Success rate\\RPM 1.2}  \\ \hline 
\thead{RL baseline \citep{Rodriguez-Ramos2019}}	& $5\si{m}\times9\si{m}$ & \thead{$\sim600\si{min}$\\$4500\si{ep}.$} & $91\%$	& $73\%$\\
						\hline 
\thead{Cascaded PI controller baseline}	& $9\si{m}\times9\si{m}$ & $-$ & $100\%$	& $100\%$\\
						\hline 
\thead{Our method }	& $9\si{m}\times9\si{m}$ & \thead{$118\si{min}$\\$2113\si{ep}.$} & $99\%$	& $99\%$\\
						\hline   	
						\hline						   								
\thead{Difference to RL baseline }	& $-$ &\thead{$\sim -80\%$\\$-53\%$} & $+8\%$	& $ +26\%$\\
						\hline
\thead{Difference to cascaded PI controller baseline }	& $-$ &$-$ & $-1\%$	& $ -1\%$\\
						\hline  			  					
\end{tabular} 
\caption{\textcolor{black}{Comparison of our approach with the baseline methods. The agent selected for the comparison is agent RPM 1.6/1. Our method significantly improves the success rate in the two reference evaluation scenarios while requiring significantly less time and fewer episodes to train compared to the RL baseline.  Furthermore, the fly zone used in our approach covers a larger area. In the reference evaluation scenarios it achieves a comparable performance than a cascaded PI controller. }}
\label{tab:comparison}
\end{table*} 

\subsection{Evaluation in Simulation with Noise}
We evaluate the selected agent of case \textit{simulation} and case \textit{hardware} \textcolor{black}{as well as the cascaded PI controller} for robustness against noise. For this purpose, we define a set of values $\sigma_{noise}= \left\lbrace  \sigma_{p_x},\sigma_{p_y}, \sigma_{p_z},\sigma_{v_x},\sigma_{v_y},\sigma_{v_z}\right\rbrace$ specifying a level of  zero mean Gaussian noise that is added to the noiseless observations in simulation. The noise level corresponds to the noise present in an EKF-based pose estimation of a multi-rotor UAV recorded during real flight experiments.

\begin{small}
\begin{align}
\sigma_{noise} &= \left\lbrace 0.1\si{m},0.1\si{m},0.1\si{m},0.25\si{m/s},0.25\si{m/s},0.25\si{m/s}\right\rbrace
\label{eq:noise_level}
\end{align}
\end{small}

We evaluate the landing performance again for static, periodic and eight-shaped trajectories of the landing platform in Table~\ref{tab:noise_evaluation_case_simulation} for training case \textit{simulation} and in Table~\ref{tab:noise_evaluation_case_hardware} for training case \textit{hardware}, \textcolor{black}{both also showing the results of the cascaded PI controller}.
For the selected agent of training case \textit{simulation}, adding the realistic noise $\sigma_{noise}$ leads to a slightly reduced performance. However, the achieved success rates are still higher than the agent's performance in  the \textcolor{black}{RL} baseline without noise. For the evaluation scenario with RPM $0.4$, our success rate is  $+4\%$ higher. With RPM $1.2$, it is $+20\%$ higher. \textcolor{black}{The cascaded PI controllers are unaffected by the specified noise level. They enable a success rate of $100\%$ across all scenarios of platform movement.}

\begin{table*}[]
\fontsize{\tfs}{\bfs}\selectfont
\centering
\begin{tabular}{|c|c|c|c|c|c|c|}
\hline
Simulation eval. $\rightarrow$   & Static & RPM  0.4 & RPM  0.8 &RPM 1.2& RPM 1.6 &8-shape \\ 
\hline
\thead{Case  \textit{simulation} \\Agent RPM 1.6/1}	& $97\%$ & $95\%$ & $98\%$ & $93\%$ & $85\%$ &$85\%$\\
\hline   
\thead{Cascaded PI \\controller}	& $100\%$ & $100\%$ & $100\%$ & $100\%$ & $100\%$ &$100\%$\\
\hline 			
\end{tabular} 
\caption{\textcolor{black}{Success rates in percent achieved in simulation with noisy observations by the best performing agent of training case \textit{simulation} and the cascaded PI controller. They have been evaluated over 150 landing trials for six different scenarios of platform movement, as is indicated by the column titles.} }
\label{tab:noise_evaluation_case_simulation}
\end{table*} 

\begin{table*}[]
\fontsize{\tfs}{\bfs}\selectfont
\centering
\begin{tabular}{|c|c|c|c|c|}
\hline
Simulation eval. $\rightarrow$   & Static & RPM  0.2 & RPM  0.4 &8-shape \\ 
\hline
\thead{Case  \textit{hardware} \\Agent RPM 0.4/3}	& $95\%$ &$91\%$ &$85\%$ & $85\%$\\
\hline   
\thead{Cascaded PI\\ controller}	& $99\%$ &$100\%$ &$99\%$ & $99\%$\\
\hline 			
\end{tabular} 
\caption{\textcolor{black}{Success rates in percent achieved in simulation with noisy observations by the selected agent of training case \textit{hardware} and the cascaded PI controller. They have been evaluated over 150 landing trials for four different scenarios of platform movement, as is indicated by the column titles.} }
\label{tab:noise_evaluation_case_hardware}
\end{table*} 

For the selected agent of training case \textit{hardware}, the drop in performance is slightly more pronounced. The reason is that the fly zone and platform specified for this training case are significantly smaller than for the case \textit{simulation}. As a consequence, the size of the discrete states is also significantly reduced, whereas the noise level stays the same. Thus, noise affects the UAV more, since it is more likely that the agent takes a suboptimal action due to an observation that was biased by noise. \textcolor{black}{The cascaded PI controller is unaffectd by noise also in the scenario of case \textit{hardware}, achieving success rates between $99\%$ and $100\%$ across the different types of platform movement.}
\textcolor{black}{
\subsection{Evaluation in Real Flight Experiments}
\subsubsection{General}
}
Unlike the baseline method, we do not evaluate our approach on real hardware in single flights only. Instead, we provide statistics on the agent's performance for different evaluation scenarios to illustrate the sim-to-real-gap.
For this purpose, the selected agent of case \textit{hardware} was deployed on  a quadcopter, see Figs.~\ref{fig:title_image} and \ref{fig:real_flights_equipment}, that has a mass of $m_{uav,RL} = 0.72 \si{kg}$  and a diameter of  $d_{uav,RL}=0.28 \si{cm}$ and deviates from the UAV used in simulation (mass $1\%$, diameter $18\%$). \textcolor{black}{The cascaded PI controller was tested on a similar quadcopter, the only difference being a different structure carrying the Vicon markers, resulting in a slightly different mass of $m_{uav,PI} = 0.696\si{kg}$. The diameter is the same.
 \begin{figure}
 \centering
  \includegraphics[width=7.5cm]{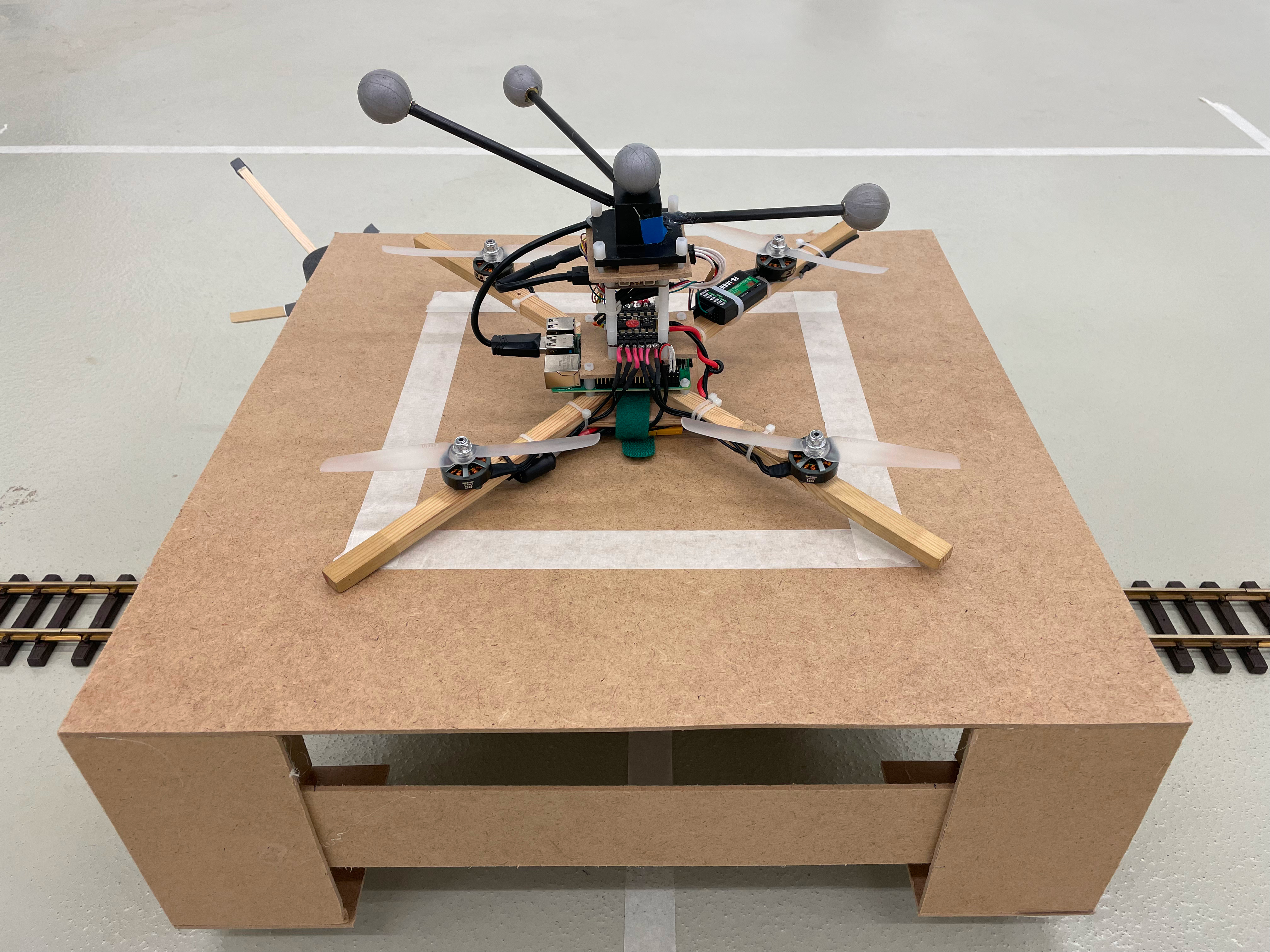}
\caption{Multi-rotor vehicle and autonomous platform moving on rails that were used for the flight experiments in the real world. The squared platform has an edge length of $0.5\si{m}$.}
\label{fig:real_flights_equipment}
\end{figure}
Both quadcopters are} equipped with a Raspberry Pi 4B providing a ROS interface and a LibrePilot Revolution flight controller to enable the tracking of the attitude angles commanded by \textcolor{black}{ the respective controllers} via ROS. \textcolor{black}{Furthermore, it also controls the commanded descend velocity.} A motion capture system (Vicon) provides almost  noiseless values of the position and velocity of the moving platform and the \textcolor{black}{quadcopter}. However, we do not use these information directly to compute the observations \eqref{eq:cont_obs_p_cx}-\eqref{eq:cont_obs_phi_rel}. The reason is that the motion capture system occasionally has wrong detections of markers. This can result in abrupt jumps in the orientation estimate of the UAV. Since the flight controller would immediately react, this could lead to a dangerous condition in our restricted indoor environment. To avoid any dangerous flight condition, we obtain the position and velocity of the UAV from an EKF-based state estimation method running on the flight controller. For this purpose, we provide the flight controller with a fake GPS signal (using Vicon) with a frequency of $10\si{hz}$. It is then fused with other noisy sensor data (accelerometer, gyroscope, magnetometer, barometer). The EKF is robust to short periods of wrong orientation estimation. \textcolor{black}{All tested algorithms are run off-board and the setpoint values for the attitude angles generated by the respective controllers are sent to the flight controller via ROS using a WLAN connection}. Due to hardware limitations regarding the moving platform, we only evaluate the agent for a static platform and the rectilinear periodic movement. 
Generally, the Vicon system allows for a state estimate of the copter that has a lower noise level than the one specified by \eqref{eq:noise_level}. However, the velocity of the platform is determined purely by means of the Vicon system. The rough surface of the ground caused vibrations of the platform that induced an unrealistic high level of noise in the Vicon system's readings of the platform velocity. For this reason, it is filtered using a low-pass filter with a cut-off frequency of $10\si{hz}$. 

 \begin{figure*}[!htb]
 \centering
  \includegraphics[scale=0.32]{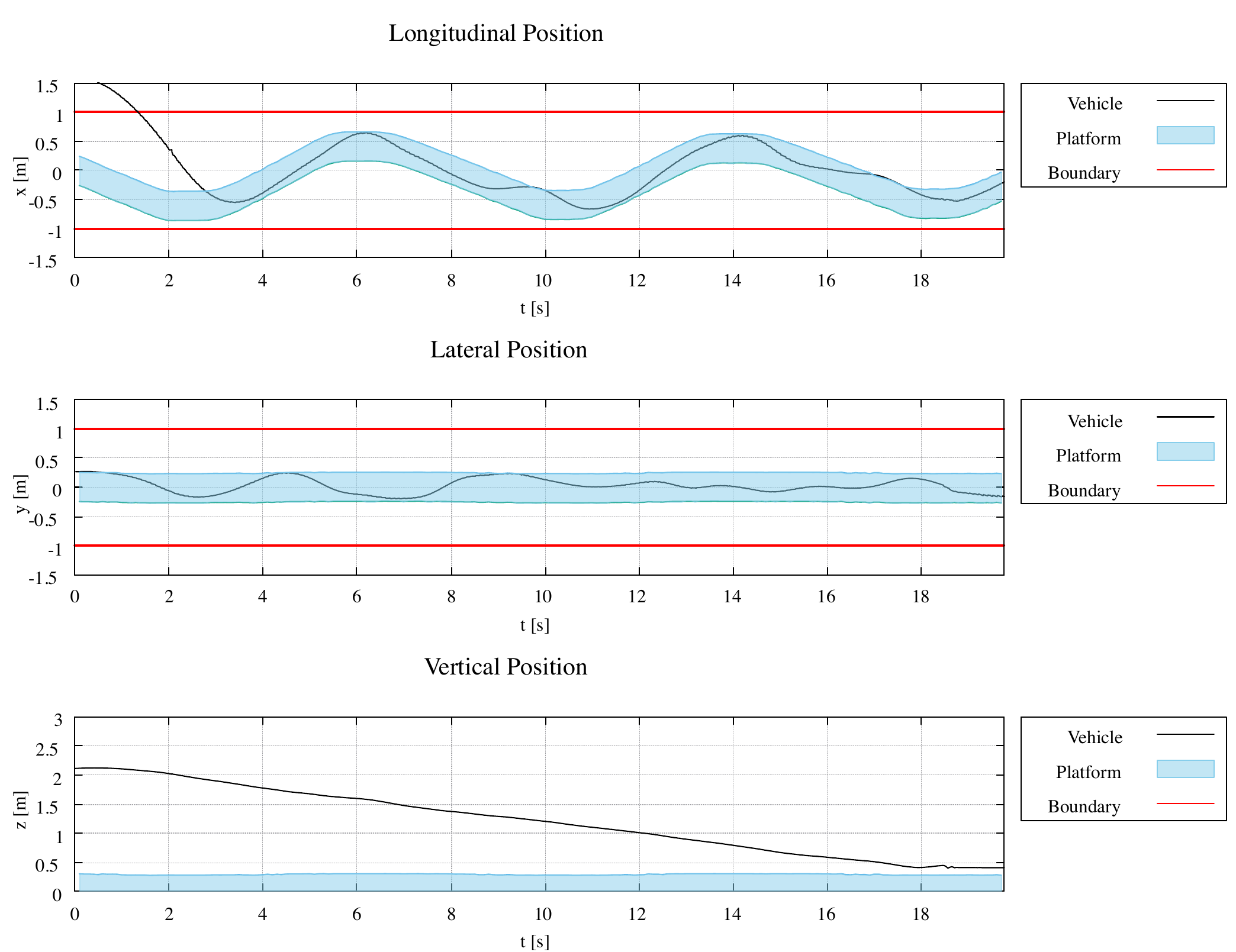}
\caption{Example trajectory of the multi-rotor vehicle and moving platform during the real flight experiment. The platform's position is determined using the Vicon system, the multi-rotor vehicle's position results from the state estimate of the flight controller. }
\label{fig:landing_trajectories_real}
\end{figure*}

\begin{table*}[!htb]
\fontsize{\tfs}{\bfs}\selectfont
\centering
\begin{tabular}{|c|c|c|c|}
\hline
Real flights eval. $\rightarrow$   & Static & RPM  0.2 & RPM  0.4 \\ 
\hline
\thead{Case  \textit{hardware} \\Agent RPM 0.4/3}	& \thead{$96\%$\\$23$ trials} &\thead{$68\%$\\$25$ trials} &\thead{$79\%$\\$28$ trials} \\
\hline   		
\end{tabular} 
\caption{\textcolor{black}{Success rates in percent achieved in real flights with ground effect by the selected agent of training case \textit{hardware}. They have been evaluated over at least 20 landing trials for three different scenarios of platform movement, as is indicated by the column titles.} }
\label{tab:real_flights_evaluation_with_ground_effect}
\end{table*} 
\textcolor{black}{
\subsubsection{Evaluation of the RL Approach}
}Fig. \ref{fig:landing_trajectories_real} shows the trajectory of the UAV and moving platform for a landing trial where the platform was executing a rectilinear periodic movement with a maximum velocity of $0.4\si{m/s}$. The depicted $x$ and $y$ component of the multi-rotor vehicle's trajectory are based on the state estimate calculated by the EKF. All other presented trajectory values are based on the readings of the Vicon system. Table~\ref{tab:real_flights_evaluation_with_ground_effect} contains the success rates achieved in the experiments \textcolor{black}{with our RL algorithm deployed on real hardware}. The starting positions of the UAV were manually selected and as uniformly distributed as possible in \textcolor{black}{an average altitude of  $2.17\si{m}$}.
Whereas for the static platform a success rate of $96\%$ could be reached, there is a noticeable drop in performance for the evaluation scenarios in which the platform is performing the rectilinear periodic movement. We argue that these can be attributed to the following main effects. 
\begin{enumerate}
\item The deviation in the size of the UAV and the different mass plays a role. \textcolor{black}{Whereas the difference in mass is seems negligible, the distribution of mass and thus the inertia of the multi-rotor vehicle is different.  }
\item A  reason for the drop in performance could also be a slightly different behaviour of the low-level controllers than in simulation. They have been tuned manually. A possible solution approach to reduce the sim-to-real gap here could be to vary the controller gains within a small range during training. This should help the agent towards better generalization properties.
\item Trials in which occasional glitches in the state estimate occurred were not treated specially although they could result in extra disturbances the RL controllers had to compensate. Furthermore, we counted also small violations of the boundary conditions as failure, such as leaving the fly zone by a marginal distance even if the agent was  able to complete the landing successfully hereafter. 
\item  The ground effect can play an important role for multi-rotor vehicles  \citep{Sanches-Cuevas2017}. Since it was not considered in the Gazebo based simulation environment used for training, it contributes to the sim-to-real gap.
\end{enumerate}
Furthermore, during the real flight experiments no significant jittering in the agent's actions could be observed. This substantiates the approach of calculating values for the  maximum pitch angle and agent frequency presented in Sect.~\ref{sec:hyperparemter_estimation}.
\textcolor{black}{
\subsubsection{Evaluation of the Cascaded PI Controller}
Table \ref{tab:evaluation_real_world_case_hardware_cascaded_pi} shows the success rates achieved with the cascaded PI controller.  Again, the starting positions of the UAV were manually selected and as uniformly distributed as possible in an average altitude of  $2.09\si{m}$. 
\begin{table*}[]
\fontsize{\tfs}{\bfs}\selectfont
\centering
\begin{tabular}{|c|c|c|c|}
\hline
Real world eval. $\rightarrow$   & Static & RPM  0.2  & RPM  0.4  \\ 
\hline
Cascaded PI	& \thead{$100\%$\\$25$ trials} & \thead{$60\%$\\$20$ trials} & \thead{$86\%$\\$14$ trials} \\
\hline   			
\end{tabular} 
\caption{\textcolor{black}{Success rates in percent achieved in real world experiments by the cascaded PI controller structure for case \textit{hardware}. It has been evaluated over different numbers of landing trials  for three different scenarios of platform movement, as is indicated by the column titles. For the cascaded PI controller, violations of the boundaries of the fly zone are not counted as failures.}}
\label{tab:evaluation_real_world_case_hardware_cascaded_pi}
\end{table*} 
The results show that there is a notable drop in performance for the scenarios RPM 0.2 and RPM 0.4. This was not expected considering the strong results obtained in simulation. Careful investigation of the log files revealed several reasons that could play a role.
\begin{enumerate}
\item The multi-rotor vehicle used for testing the cascaded PI controller differed from the one used for the evaluation of the RL controller. Although being identical in terms of size and components, it is $0.02\si{kg}$ lighter due to a different structure carrying the markers for the Vicon system.
\item In $81\%$ of the landing trials a loss of state estimate messages that were sent by the multi-rotor vehicle and received by the ground station running the cascaded PI controller occurred. This happened probably due to issues of ROS queing messages or limited bandwidth of the WLAN module, which was changed. The cascaded PI controller is intended and tuned to run at $100\si{hz}$, like in the Gazebo simulation, however, had to handle low frequencies of approximately $60\si{hz}$ instead for different lengths of time. As a consequence,  the cascaded PI controller's cycles were executed with a non-constant and often reduced frequency. However, drops in the frequency of the received state estimate were also present for the scenario of a static platform where the cascaded PI controller was able to achieve high success rates. The necessity to transmit attitude setpoints with high frequency can be a disadvantage in situations in which the control method is run off-board and the resulting commands are sent to the multi-rotor vehicle via WLAN. In this regard, our RL based approach is beneficial as it is designed to be run at considerably lower frequencies as is illustrated by Table \ref{tab:training_differences}. Drops in frequency could also be noticed in $58\%$ of the landing trials during the evaluation of the  RL controller.  
\item All flight experiments were conducted inside a virtual fly box where the tested controller is given authority to control the multi-rotor vehicle. For the RL controller the virtual fly box was set to $[-2\si{m},2.2\si{m}]\times[-1\si{m},1\si{m}]\times[0.4\si{m},2.6\si{m}]$ and for the cascaded PI controller to $[-1\si{m},1\si{m}]\times[-1\si{m},1\si{m}]\times[0\si{m},2.3\si{m}]$. As soon as the virtual fly box is left a fallback controller steers the multi-rotor vehicle back inside before control is handed over to the tested controller again. This maneuver can increase the momentum of the copter which then has to be handled by the limited attitude angles of the tested controller. Leaving the virtual fly box happened frequently during testing of the cascaded PI controller where the initial position of a landing trial was often located outside the fly box, probably due to issues with the altitude estimation. For this reason, leaving the fly box is not counted as a failed attempt for the cascaded PI controller in Table \ref{tab:evaluation_real_world_case_hardware_cascaded_pi}.
\end{enumerate}
Besides the aforementioned issues, an additional problem of the cascaded PI controller became obvious during the real world experiments. Without a differential branch it cannot react to the current change in the control errors of the outer and inner cascade. During tuning in simulation a cascaded PI controller turned out to be successful. However, during the real world experiments it could be noted that the cascaded PI controller tends to overshoot the platform in the scenarios RPM 0.2 and RPM 0.4. Here, the RL controller is advantageous because it has learned how to handle the platform's change in the direction of motion.
}

\textcolor{black}{
\subsection{Ablation Study}
In the context  of our approach and in addition to the presented method where the landing task was learned by means of a sequential curriculum leveraging transfer learning between curriculum steps, there are two other ways of how learning the task could be achieved. First,  the sequential curriculum is conducted without transfer learning. This means that whenever a new curriculum step is added to the learning sequence the Q-table associated with the latest curriculum step is initialized with zeros. We denote this ablation case \textit{Sequential curriculum without transfer learning}. And second, we abandon curriculum learning completely. Instead, we create the full sequence of learning steps right in the beginning with all associated Q-tables initialized with zeros. We then conduct one training on the full sequence without adding an additional step after each round of training. This constitutes a scenario in which  several separate Q-learning tasks are trained simultaneously. For this reason, we call this ablation case \textit{Simultaneous training of sequence}. For the training and evaluation, we chose the scenario RPM 1.6. All learning related parameters are the same as presented in Sect. \ref{sec:training}. We compare these additional two methods with the agent selected for further evaluation in Sect. \ref{sec:agent_selection} and name this scenario \textit{Sequential curriculum with transfer learning}. Table \ref{tab:ablation_study} summarizes the results.}\\
\begin{table*}[thpb]
\fontsize{\tfs}{\bfs}\selectfont
\centering
\begin{tabular}{|c|c|c|c|c|c|c|c|c|}
\hline
Ablation case  & Static $[\%]$ &  \thead{\fontsize{\tfs}{\bfs}\selectfont RPM\\ \fontsize{\tfs}{\bfs}\selectfont 0.4 $[\%]$} &  \thead{\fontsize{\tfs}{\bfs}\selectfont RPM\\ \fontsize{\tfs}{\bfs}\selectfont 0.8 $[\%]$}  &  \thead{\fontsize{\tfs}{\bfs}\selectfont RPM\\ \fontsize{\tfs}{\bfs}\selectfont 1.2 $[\%]$} & \thead{\fontsize{\tfs}{\bfs}\selectfont RPM\\ \fontsize{\tfs}{\bfs}\selectfont 1.6 $[\%]$} & \thead{\fontsize{\tfs}{\bfs}\selectfont 8-shape $[\%]$}&\thead{\fontsize{\tfs}{\bfs}\selectfont Training\\\fontsize{\tfs}{\bfs}\selectfont duration $[min]/$\\\fontsize{\tfs}{\bfs}\selectfont number of episodes  $[-]$}\\ 
\hline
\thead{\fontsize{\tfs}{\bfs}\selectfont{\fontsize{\tfs}{\bfs}\selectfont Sequential curriculum}\\ {\fontsize{\tfs}{\bfs}\selectfont without transfer}\\ {\fontsize{\tfs}{\bfs}\selectfont learning}}	& $57$ & $60$  & $59$ & $46$ & $42$ & $43$ & \thead{\fontsize{\tfs}{\bfs}\selectfont $520$\\\fontsize{\tfs}{\bfs}\selectfont $5767$}\\
\hline   						
\thead{{\fontsize{\tfs}{\bfs}\selectfont Simultaneous training}\\ {\fontsize{\tfs}{\bfs}\selectfont of sequence}}	& $81$ & $86$  & $77$ & $74$ & $72$ & $72$ & \thead{\fontsize{\tfs}{\bfs}\selectfont $147$\\\fontsize{\tfs}{\bfs}\selectfont $1538$}\\
\hline   						
\thead{{\fontsize{\tfs}{\bfs}\selectfont Sequential curriculum}\\ {\fontsize{\tfs}{\bfs}\selectfont with transfer}\\ {\fontsize{\tfs}{\bfs}\selectfont learning}}	& $99$ & $99$  & $98$ & $99$ & $93$ & $93$& \thead{\fontsize{\tfs}{\bfs}\selectfont $118$\\\fontsize{\tfs}{\bfs}\selectfont $2113$}\\
\hline   						
\end{tabular} 
\caption{\textcolor{black}{Results of the ablation study. The columns two to seven show the success rates achieved in the different evaluation scenarios. The last column contains the training information of the respective ablation case. }}
\label{tab:ablation_study}
\end{table*} 
\textcolor{black}{It is clear that ablation case \textit{Sequential curriculum without transfer learning} performs worst across all evaluation scenarios in terms of both training duration as well as number of episodes required. Our rationale here is the following. In each new curriculum step the task has to be learned from scratch. However, since in all curriculum steps added to the sequence after the initial step the exploration rate is set to $\varepsilon=0$, it is even more difficult to learn optimal behaviour. Furthermore, in the early phase after adding a new curriculum step its behaviour is  bad due to few updates of the state-action-pairs, therefore acting as a source of disturbance to the previous curriculum steps. For this reason, in order to reach a  situation in which the Q-values of the latest curriculum step as well as the Q-values of the previous curriculum steps have sufficiently adapted ($96\%$ success rate over the last $100$ episodes), a significant amount of time is required. The discrepancy between the success rate necessary to end the training and the success rate achieved during evaluation can be explained as follows. As outlined in Sect. \ref{sec:curriculum_discretization}, an episode is terminated with success if the agent spends one second in the latest curriculum step before reaching its goal state. However, during evaluation the multi-rotor vehicle is required to constantly stay above the moving platform, not only for one second. This leaves room for  erratic behaviour after this one second period which can be observed for this ablation case. Even though centering the multi-rotor vehicle above the platform is learned in principle, maintaining that centered state for a longer time period is not possible which  leads to low success rates. Conceivably, the situation could be improved by choosing a longer time period for maintaining the goal state before the success criterion is met. \\
Allowing training on all curriculum steps that were initialized with zeros simultaneously creates the possibility to adapt all Q-values of all the separate Q-tables associated with the different steps in the sequence during one training. This solves the problem where the latest untrained curriculum step acts as a disturbance to the previous curriculum steps. As a result, the training time is considerably shorter for the ablation case \textit{Simultaneous training of sequence}. Also, the erratic behaviour is reduced, leading to higher success rates across all evaluation scenarios. However, at the end of the training the number of unvisited state-action-pairs is quite high, $497$ out of $2835$ state-action pairs. This indicates that the training is not yet optimal, although the terminal criterion to end the training has been met. This might be solved with a prolongation of the exploration phase.\\
The best performance in terms of success rates as well as required training time was achieved by the third ablation case \textit{Sequential curriculum with transfer learning} which constitutes the full method. Here, the erratic behaviour is not present anymore. We hypothize that this is due to the transfer of knowledge between curriculum steps. Not only does this significantly reduce the number of unvisited state-action-pairs ($0$ out of $2385$, a curriculum step adds $567$ new state-action-pairs to the sequence) but also the knowledge does not have to be learned from scratch in each new curriculum step but is rather only refined.}

\section{Conclusion and Future Work}\label{sec:conclusion}
In this work, we presented a RL based method for autonomous multi-rotor landing. Our method splits the overall task into simpler 1-D tasks, then formulates each of those as a sequential curriculum  in combination with a knowledge transfer between curriculum steps. Through rigorous experiments, we demonstrate significantly shorter training time ($\sim -80\%$) and higher success rates (up to $+26\%$) than the DRL-based actor-critic method presented in \citep{Rodriguez-Ramos2019}. \textcolor{black}{For two comparison scenarios, our method achieves  a performance comparable to a cascaded PI controller.} We present statistics of the performance of our approach on real hardware and show interpretable ways to set hyperparameters. \textcolor{black}{In the future, we plan to test other tabular RL methods such as n-step $Q(\sigma)$ learning or the use of eligible traces. Furthermore, extending the approach to control the vertical movement and yaw angle of the multi-rotor vehicle is of interest as well as addressing complex landing problems, such as on an inclined or on a flying platform}.

\subsubsection*{Acknowledgements} 
The authors thank Eric Price for his extensive support with the flight experiments. \\
The authors thank the International Max Planck Research School for Intelligent Systems (IMPRS-IS) for supporting Pascal Goldschmid.

\subsubsection*{Funding}
This work was funded by the Institute of Flight Mechanics and Controls, University of Stuttgart.

\subsubsection*{Author Contributions}
Pascal Goldschmid is the first author, implemented the methods, performed all experiments and prepared the main manuscript. Aamir Ahmad supervised the work, provided guidance to the first author, and revised and edited the manuscript. Both authors reviewed the paper.

\subsubsection*{Conflict of Interest}
Both authors' primary affiliation is the Institute of Flight Mechanics and Controls, University of Stuttgart.
Pascal Goldschmid is employed as a research associate and Aamir Ahmad as a tenure-track professor. Both authors  are also affiliated with the Perceiving Systems department of the Max Planck Institute for Intelligent Systems.